\pdfoutput=1

\documentclass[11pt]{article}

\usepackage[acceptedWithA]{tacl2021v1}

\usepackage{times}
\usepackage{latexsym}
\usepackage{todonotes}

\usepackage[T1]{fontenc}

\usepackage[utf8]{inputenc}

\usepackage{microtype}

\usepackage{inconsolata}

\usepackage{graphicx}
\usepackage{multirow}
\usepackage{float}
\usepackage{booktabs}
\usepackage{amssymb}
\usepackage{pifont}
\usepackage{caption}
\usepackage{subcaption}
\usepackage{enumitem}

\newcommand{\cmark}{\ding{51}}
\newcommand{\xmark}{\ding{55}}

\newcommand{\clapnq}{CLAP\textsc{nq}}
\newcommand{\fiqa}{FiQA}
\newcommand{\govt}{Govt}
\newcommand{\cloud}{Cloud}

\newcommand{\ragasF}{RL$_\textrm{F}$}
\newcommand{\ragasR}{RL$_\textrm{R}$}
\newcommand{\radbench}{$\textrm{RB}_\textrm{llm}$}
\newcommand{\agg}{$\textrm{RB}_\textrm{alg}$}
\newcommand{\mtb}{RL$_\textrm{MTB}$}

\newcommand{\rougeL}{Rouge-L}
\newcommand{\bkPrec}{Bert-K-Prec}
\newcommand{\bRec}{Bert-Rec}

\newcommand{\fanc}{\textit{FANC}}

\newcommand\sbullet[1][.5]{\mathbin{\vcenter{\hbox{\scalebox{#1}{$\bullet$}}}}}

\newcommand{\benchmark}{\textsc{mt}{RAG}}
\newcommand{\benchmarksynth}{\textsc{mt}{RAG-S}}

\newcommand*\halfcirc[1][1ex]{%
  \begin{tikzpicture}
  \draw[fill] (0,0)-- (90:#1) arc (90:270:#1) -- cycle ;
  \draw (0,0) circle (#1);
  \end{tikzpicture}}

\title{\benchmark: A Multi-Turn Conversational Benchmark for\\ Evaluating Retrieval-Augmented Generation Systems}

\author{
\textbf{Yannis Katsis$^*$, Sara Rosenthal$^*$, Kshitij Fadnis, Chulaka Gunasekara,
Young-Suk Lee,}\\
\textbf{Lucian Popa, Vraj Shah, Huaiyu Zhu, Danish Contractor, Marina Danilevsky}\vspace{0.2cm}\\
IBM Research\vspace{0.2cm}\\
\texttt{\small yannis.katsis@ibm.com, 
sjrosenthal@us.ibm.com,
kpfadnis@us.ibm.com,}\\
\texttt{\small chulaka.gunasekara@ibm.com,
ysuklee@us.ibm.com,
lpopa@us.ibm.com, 
vraj@ibm.com,}\\ 
\texttt{\small huaiyu@us.ibm.com, 
Danish.Contractor@ibm.com,
mdanile@us.ibm.com}
}

\begin{document}
\maketitle

\def\thefootnote{*}\footnotetext{These authors contributed equally to this work.}\def\thefootnote{\arabic{footnote}}

\begin{abstract}
Retrieval-augmented generation (RAG) has recently become a very popular task for Large Language Models (LLMs). Evaluating them on \emph{multi-turn} RAG conversations, where the system is asked to generate a response to a question in the context of a preceding conversation is an important and often overlooked task with several additional challenges. We present \benchmark: an end-to-end human-generated multi-turn RAG benchmark that reflects several real-world properties across diverse dimensions for evaluating the full RAG pipeline.  \benchmark\ contains 110 conversations averaging 7.7 turns each across four domains for a total of 842 tasks. We also explore automation paths via synthetic data and LLM-as-a-Judge evaluation. Our human and automatic evaluations show that even state-of-the-art LLM RAG systems struggle on \benchmark. We demonstrate the need for strong retrieval and generation systems that can handle later turns, unanswerable questions, non-standalone questions, and multiple domains.
\benchmark\ is available at \url{https://github.com/ibm/mt-rag-benchmark}.
\end{abstract}

\section{Introduction}
\label{sec:intro}

\begin{figure}[t]
    \centering
    \includegraphics[width=\columnwidth]{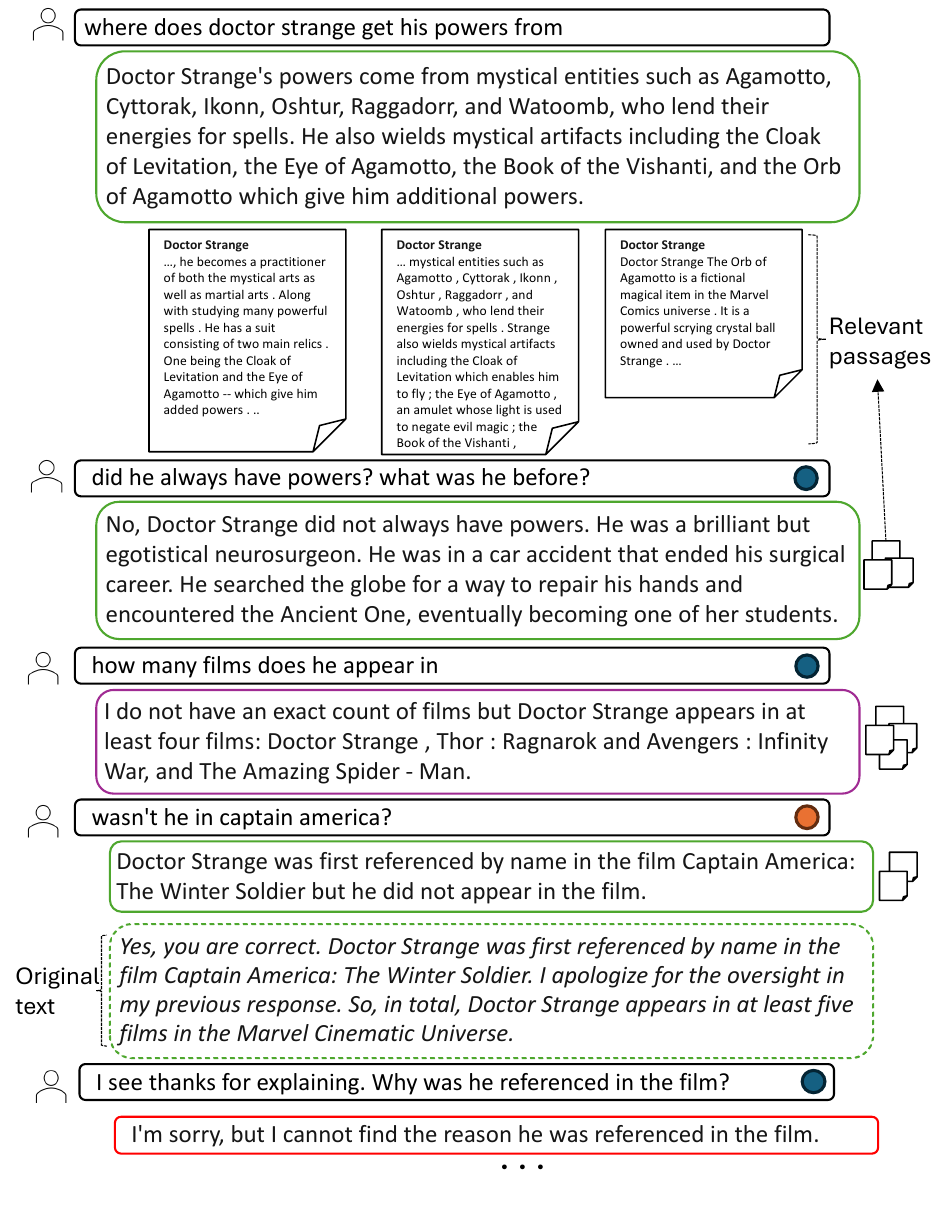}
    \caption{\small 5/8 turns of a conversation from the \clapnq\ domain. The conversation is enriched with question dimensions, passage diversity, and repair. The answerability is shown using the  agent response color: 
    \textcolor{green}{answerable}, \textcolor{red}{unanswerable}, and \textcolor{purple}{partial}. 
    The multi-turn type is shown using the question circle: \textcolor{blue}{follow-up} and \textcolor{orange}{clarification}. The different relevant passages highlight diversity and the original text shows a repair from the model response.}
    \label{fig:conversation_example}
\end{figure}

Large Language Models (LLMs) play an important role as chat-based assistants \cite{lin2023review}.  Relying on knowledge-sources during the conversation is an important task that helps improve answer reliability and trust, and hence Retrieval-augmented generation (RAG) has become an important and popular field in recent years \cite{lewis2020retrieval, chen2024benchmarking}. The primary focus of RAG benchmarks has been on single turn \cite{friel2024ragbenchexplainablebenchmarkretrievalaugmented, niu2023ragtruth, yang2024crag} which LLMs have become proficient at \cite{kuo2024radbenchevaluatinglargelanguage}, however multi-turn RAG, where a turn is defined as a question-response pair, has been largely overlooked as a benchmark, and presents additional challenges not covered in single-turn RAG. A multi-turn conversation benchmark should sufficiently cover several challenging aspects to effect a holistic evaluation of the  full RAG pipeline:

\begin{description}[style=unboxed,leftmargin=0cm]
   \item[Retrieval] The relevant passages should change during the conversation causing repeated retrieval. 
    \item[Generation] The generator should struggle to answer many of the questions correctly, particularly questions that refer to and rely on previous turns.
\end{description}

We present \benchmark, a diverse and representative multi-turn RAG benchmark of human-generated conversations across 4 different domains that vary in style, topic and source. Our conversations comprise turns that vary along the dimensions of \textit{question type}, \textit{multi-turn}, and \textit{answerability}.

Our benchmark is constructed using a novel process where human annotators simulate a real-world conversation, by actually interacting with a live RAG agent via a custom chat application and improving the output in real time. 
Annotators took care to diversify their questions across the aforementioned different dimensions, including referencing earlier turns, while ensuring a flowing and natural conversation. At every turn, after issuing their questions, annotators checked the passages retrieved by the RAG system and modified the passage set to improve relevance and diversity. Next, they reviewed and repaired the generated response to improve its quality. Figure~\ref{fig:conversation_example} shows part of a conversation from the benchmark to illustrate the output of our data creation process (described in detail in Section~\ref{sec:dataset_creation}). The resulting conversations average 7-8 turns in length and 16.9 unique relevant passages per conversation. 

We evaluate our \benchmark{} benchmark on the retrieval and generation components of RAG systems. We examine retrieval performance of lexical, sparse and dense retrieval under two settings (last turn and query rewrite) and analyze generative performance of 9 LLMs under three retrieval settings (reference, reference+RAG, and full RAG). Our rigorous human evaluation of model responses demonstrates that all models struggle on our tasks, especially on unanswerable questions, and in later turns.

As human data generation and evaluation does not scale well, we systematically explore automation paths. We identify several automated evaluation metrics that correlate with human scores (and many that do not), and demonstrate the need for more work in automatic evaluation. We also construct a companion benchmark, \benchmarksynth, of synthetically generated conversations. By providing both human-generated and synthetic conversations over the same corpora, we aim to help the community analyze and understand the relative advantages of the two types of data.

Our contributions are as follows:

\begin{enumerate}
    \item We present \benchmark, a comprehensive and diverse human-generated multi-turn RAG benchmark, accompanied by four document corpora.
    The benchmark is available at \url{https://github.com/ibm/mt-rag-benchmark}.
    \item We evaluate lexical, dense and sparse retrieval models and 9 large language models and conduct a rigorous human evaluation of model responses. 
    \item We systematically explore automation paths for data and metrics, demonstrating gaps in current automatic evaluation and constructing a companion benchmark, \benchmarksynth, of synthetically generated conversations.
\end{enumerate}

To the best of our knowledge, \benchmark{} is the first end-to-end human-generated multi-turn RAG benchmark that reflects real-world properties of multi-turn conversations.

\begin{table}[t]
\centering
\small
\setlength{\tabcolsep}{2pt}
\begin{tabular}{l|c|c|c|c}
\toprule
\multicolumn{1}{c|}{Benchmark}                                           & \begin{tabular}[c]{@{}c@{}}\scriptsize{Active} \\ \scriptsize{Retriev.}\end{tabular} & \begin{tabular}[c]{@{}c@{}}\scriptsize{Long} \\ \scriptsize{Answer}\end{tabular} & \begin{tabular}[c]{@{}c@{}}\scriptsize{Unans-}\\ \scriptsize{werable}\end{tabular} & \begin{tabular}[c]{@{}c@{}}\scriptsize{Multi-}\\ \scriptsize{Domain}\end{tabular}                                                                                       \\ \midrule
QuAC\scriptsize{\cite{choi-etal-2018-quac}}                        & \xmark                                                                 & \xmark                                       & \cmark                                          & \xmark                                                                                     \\
Or-QuAC\scriptsize{\cite{qu2020open}}                               & \xmark                                                                 & \xmark                                       & \cmark                                          & \xmark                                                                                     \\
CoQA\scriptsize{\cite{reddy-etal-2019-coqa}}                        & \xmark                                                                 & \xmark                                       & \cmark                                          & \xmark                                                                                     \\
ShARC\scriptsize{\cite{saeidi-etal-2018-interpretation}}            & \xmark                                                                 & \xmark                                       & \xmark                                          & \xmark                                                                                         \\
MD2Dial\scriptsize{\cite{feng2021multidoc2dial}}              & \xmark                                                                 & \xmark                                       & \xmark                                          & \xmark                                                                                         \\
FaithDial\scriptsize{\cite{dziri-etal-2022-faithdial}}              & \cmark                                                                 & \xmark                                       & \cmark                 & \xmark                                                                                     \\
iKAT\scriptsize{\cite{10.1145/3626772.3657860}} & \xmark & \cmark & \xmark & \xmark \\
RADBench\scriptsize{\cite{kuo2024radbenchevaluatinglargelanguage}} & \xmark                                                                 & \cmark                                       & \xmark                                          & \cmark  \\ \midrule
{\bf \benchmark\ (Our work)}                       & \cmark                                                                 & \cmark                                       & \cmark                                          & \cmark \\ \bottomrule
\end{tabular}
\caption{\small \benchmark\ compared to prior multi-turn RAG benchmarks}
\label{tbl:benchmark-comparison}
\end{table}

\section{Related Work}
\label{sec:related}

Question Answering (QA) and Information Retrieval (IR) have been popular tasks for many years. Prior focus has been on Extractive QA \cite{rajpurkar-etal-2016-squad, kwiatkowski-etal-2019-natural}, Long Form Question Answering \cite{fan-etal-2019-eli5, stelmakh-etal-2022-asqa, rosenthal2024clapnq}, and Open Domain QA \cite{lee-etal-2019-latent-opennq}. There has also been work on multi-turn conversations, primarily through datasets like MT-Bench \cite{zheng-mt-bench-2024}, which focuses on tasks without retrieval, and Wizard of Wikipedia \cite{dinan2018wizard}, which is conversational rather than information seeking. Several surveys exist which summarize prior approaches and datasets available to the community \cite{morethanreadingcomprehension2021, reviewofpublicdatasetsinqa2021, wang2022modernquestionansweringdatasets, qadatasetexplosion2023}. 

We compare our benchmark to existing multi-turn RAG datasets in Table \ref{tbl:benchmark-comparison}. Except for FaithDial \cite{dziri-etal-2022-faithdial}, prior work keeps the retrieval component fixed. Retrieval is either performed once at the beginning, restricting the entire conversation to the initial passages; or retrieval is only deployed to find evidence for an existing conversation, which can result in strange mismatches between  answers and verifiability. In contrast, we perform \emph{active retrieval} \cite{jiang-etal-2023-active}, where ongoing passage retrieval influences both follow-up questions and provided answers, more closely reflecting real life scenarios.
Excluding RAD-Bench \cite{kuo2024radbenchevaluatinglargelanguage} and iKAT \cite{10.1145/3626772.3657860}, most prior datasets focus on extractive or short answers (1-2 sentences) limiting the kind of questions that can be asked. Further, many existing datasets ignore unanswerable questions - a ripe source of hallucinations in LLMs~\cite{song2024measuringenhancingtrustworthinessllms}. Finally, most datasets focus on a single domain/topic, while we explore several domains of different types.
Our \benchmark\ benchmark reflects real-world properties of multi-turn conversations, including active retrieval, long-form answers, unanswerable questions, and multiple domains. 
\section{\benchmark\ Benchmark}
\label{sec:creation}

We now describe \benchmark's characteristics, before delving into how it was created in Section \ref{sec:dataset_creation}.

\subsection{Dimensions}
\label{sec:dimensions}

To ensure that \benchmark\ is representative of real RAG use cases, we designed it to be diverse across several important dimensions. For the detailed definitions see Appendix \ref{sec:diff-types-def}. 

\noindent{\bf Question types:} Similar to contemporaneous work on datasets for RAG \cite{yang2024crag, leegunasekara2024} conversations in \benchmark\ contain questions of diverse types, including {\em factoid}, {\em comparison}, {\em explanation}, {\em keyword} questions, and others. Each question has one or more question type labels.

\noindent{\bf Multi-turn:} In addition to question types that apply to individual questions regardless of the surrounding conversation, there are different types of multi-turn questions. 
A question  can be a \textit{follow-up } or \textit{clarification}.
All questions in \benchmark\ beyond the first turn are labeled with one multi-turn type.

\noindent{\bf Answerability:} 
Models often struggle when confronted with questions or problems that cannot be answered \cite{rajpurkar-etal-2018-know, sun2024benchmarkinghallucinationlargelanguage}. 
Based on this observation, we include \emph{answerable}, \emph{partially answerable}, \emph{unanswerable} based on the corpora, and \emph{conversational} statements (e.g., ``Hi", ``That's interesting", ``Thank you"). 

\noindent{\bf Domain:}
To test RAG systems over different types of documents, \benchmark\ is created over four domains and corresponding document corpora (discussed in more detail in Section \ref{sec:corpora}).

\begin{table*}[t]
\centering
\small
\begin{tabular}{l l l r r r}
    \toprule
    {\bf Corpus} & {\bf Domain} & {\bf Provenance} & {\bf Documents (D)} & {\bf Passages (P)} & {\bf Avg P/D}\\
    \midrule
    \clapnq & Wikipedia & \cite{rosenthal2024clapnq} & 4,293 & 183,408 & 42.7\\
    \fiqa & Finance & \cite{macedo2018fiqa} & 57,638* & 61,022 & 1.1\\
    \govt & Government & New corpus & 7,661 & 49,607 & 6.5\\
    \cloud & Technical documentation & New corpus & 8,578 & 72,442 & 8.4\\
    \bottomrule
\end{tabular}
\caption{\small Statistics of document corpora included in the \benchmark\ benchmark. (*) For \fiqa\ we report the number of individual forum posts, as the dataset does not have the notion of a document}
\label{tbl:corpora-stats}
\end{table*}

\subsection{Conversation Properties}
\label{sec:other-characteristics}

To ensure that \benchmark\ is a challenging benchmark,
the following properties were incorporated during conversation creation.

    \noindent{\bf High Quality Responses:} Following and extending prior work \cite{rosenthal2024clapnq, es-etal-2024-ragas, 10.1145/3626772.3657860}, each reference answer is written to satisfy the following properties: (a) \emph{Faithfulness:} The answer is faithful to the passages or earlier turns, (b) \emph{Appropriateness:} It is appropriate/relevant to the question, (c) \emph{Naturalness:} The answer sounds natural,  and (d) 
    \emph{Completeness:} Includes all information in the passages relevant to the question. We refer to these properties as \fanc.
    
    \noindent{\bf Passage Diversity:} Conversations contain questions that are diverse enough that the relevant passages do not remain static through the conversation. Our conversations have on average 16.9 unique relevant passages and 20.9 relevant passages in total.
    
    \noindent{\bf Answer Repair:} During conversation creation we employ an LLM to create the initial response, which is then repaired by the annotator as needed. A response that does not require repair can be considered an indication that the question is not challenging for the LLM. Our conversations contain repairs on 92\% of the turns.
    
    \noindent{\bf Non-standalone:} The questions occasionally rely on prior turns in the conversation (e.g., by using co-references to prior questions or answers or by clarifying previous turns). On average 1.3 questions per conversation include co-references.

\section{Benchmark Creation}
\label{sec:dataset_creation}

\subsection{Annotators}

The annotators that contributed to this work are highly skilled individuals hired solely to perform language annotation tasks and paid well above minimum wage. Unless otherwise noted, the annotators were used for all annotation tasks in this paper. Great care was taken to ensure random assignment.

\subsection{Document Corpora}
\label{sec:corpora}

The first step in creating the benchmark was to assemble the document corpora over which the conversations would be built. \benchmark\ consists of four document corpora/domains:

{\bf \clapnq} \cite{rosenthal2024clapnq}: a subset of Wikipedia pages,

{\bf \fiqa} \cite{macedo2018fiqa}: a set of StackExchange posts discussing financial advice,

{\bf \govt}: the crawled contents of select web-pages under the .gov and .mil domains, and

{\bf \cloud}: the crawled contents of select technical documentation pages of a major cloud provider.

\clapnq\ and \fiqa\ are existing corpora from QA/IR datasets, while \govt\ and \cloud\ are new corpora assembled specifically for this benchmark. 
To ensure that the new corpora are well suited for generating diverse conversations (i.e., conversations that touch several different passages), both the Govt and Cloud corpora were designed to contain sets of inter-connected pages (see Appendix~\ref{sec:corpora-details}).

\interfootnotelinepenalty=200
 
Each corpus was indexed using Elser from ElasticSearch.\footnote{\href{https://www.elastic.co/guide/en/machine-learning/current/ml-nlp-elser.html}{https://www.elastic.co/guide/en/machine-learning/current/ml-nlp-elser.html}} During ingestion, documents were split into passages of 512 tokens with an overlap stride of 100 tokens. The indexes are used to perform retrieval during conversation creation and all RAG experiments.
Table \ref{tbl:corpora-stats} summarizes the corpora.
 
Finally, for each corpus we also assembled a set of seed questions to help human annotators bootstrap the conversation generation process, described in Section~\ref{sec:human-gen-conversations}. For \clapnq\ and \fiqa\ we leveraged questions from the corresponding QA datasets, while for \govt\ and \cloud\ we selected a set of seed documents from each corpus and asked annotators to write seed questions based on them.

\begin{figure*}[t]
    \centering
    \includegraphics[width=\textwidth]{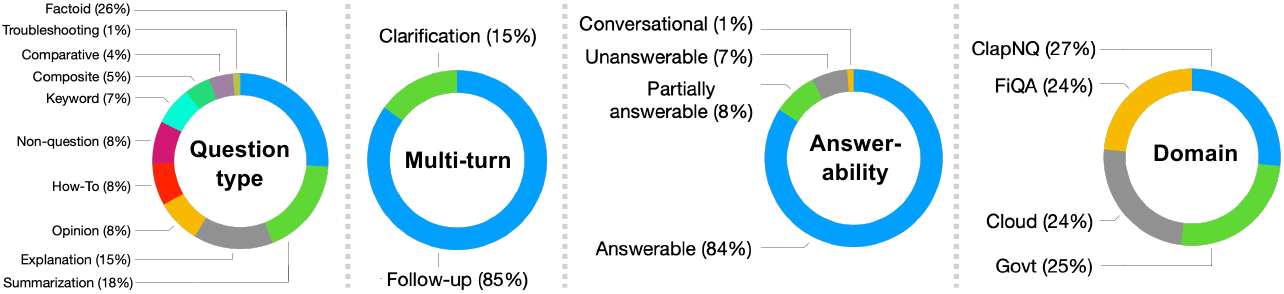}
\caption{\small Distribution of tasks in \benchmark\ based on each of the benchmark's dimensions.}
\label{fig:combined-stats-by-dimension}
\end{figure*}

\subsection{Human-Generated Conversations}
\label{sec:human-gen-conversations}

Annotators were asked to create multi-turn conversations over the four corpora. To aid in this process, we developed a custom chat application allowing them to interact with a live RAG agent consisting of an ELSERv1 (ElasticSearch 8.10)$^1$ retriever and Mixtral 8X7b Instruct \cite{jiang2024mixtralexperts} generator and correcting the retriever and generator outputs as needed. Using the application, annotators created multi-turn conversations by performing the following actions at every turn: write a question, adjust the set of retrieved passages, edit agent response and enrich with question dimensions. Annotators were directed to write 6-9 (or more) turns per conversation.

 \noindent (i) {\bf Training:} Prior to the main creation task we provided annotators with interactive training sessions, extensive documentation with examples, and pilot tasks with feedback. Once they were comfortable with the task and producing conversations that met our criteria, we began the full annotation task.

 \noindent (ii) {\bf Creation:} Annotators started a conversation using a seed question.
 For subsequent turns, they were encouraged to create questions that naturally extended the preceding conversation while varying in answerability types, question types, and multi-turn patterns (Section \ref{sec:dimensions}), which were noted during annotation time. Once annotators wrote a question, the chat application queried the retriever for potentially relevant passages, ran the generator to produce an agent response based on the retrieved passages, and presented everything to the user for repair. Annotators were guided to ensure passage diversity and answer quality (Section~\ref{sec:other-characteristics}) using the application's custom passage search and edit agent response functionalities. In the final benchmark, the similarity between the original and repaired response based on \rougeL\ (denoted as the \emph{edit score}) is 60.7 indicating significant amount of repair. This novel process simulates real-time conversations, which is missing in prior work (Section~\ref{sec:related}). 

 \noindent (iii) {\bf Review:} The resulting 126 conversations were then reviewed. Annotators could accept or reject conversations, and repair responses, passage relevance, and dimensions as needed. They were not allowed to edit the questions or passages as such changes could negatively affect the conversation flow. During this phase, most conversations were kept (see review process details in Appendix~\ref{sec:conv-review}). Our human evaluation (Section \ref{sec:human-eval}) shows that the reference responses are preferred by humans.

\subsection{Data Statistics}

This process yielded a benchmark of 110 conversations (29 ClapNQ, 27 FiQA, 28 Govt, 26 Cloud) with an average 7.7 turns per conversation, leading to 842 tasks. A task is a conversation turn containing all previous turns together with the last user question (e.g., the task created for turn $k$ includes all user and agent questions/responses for the first $k-1$ turns plus the user question for turn $k$). All evaluations described next are performed at the task level. Figure \ref{fig:combined-stats-by-dimension} shows the distribution of tasks in \benchmark\ on the benchmark dimensions. 

\section{Retrieval}
\label{sec:retrieval}

\begin{table*}
\small
\centering
\begin{tabular}{c||l|l|l|l|l||l|l|l|l}
\toprule
&	& \multicolumn{4}{c||}{\bf Recall}			&	\multicolumn{4}{c}{\bf nDCG} \\
	&	&	@1	&	@3	&	@5	&	@10	&	@1	&	@3	&	@5	&	@10	\\
 \midrule
	BM25	&	Last Turn	&	0.08	&	0.15	&	0.20	&	0.27	&	0.17	&	0.16	&	0.18	&	0.21	\\
	&	Query Rewrite	&	\textbf{0.09}	&	\textbf{0.18}	&	\textbf{0.25}	&	\textbf{0.33}	&	\textbf{0.20}	&	\textbf{0.19}	&	\textbf{0.22}	&	\textbf{0.25}	\\
 \midrule
 BGE-base 1.5 &	Last Turn	&	0.13	&0.24	&0.30&	0.38 &0.26	&0.25	&0.27	&0.30\\
&	Query Rewrite	&\textbf{0.17}&	\textbf{0.30}&	\textbf{0.37}&	\textbf{0.47} &\textbf{0.34}	&\textbf{0.3}1&	\textbf{0.34}	&\textbf{0.38}\\
 \midrule
Elser	&	Last Turn	&	0.18	&	0.39	&	0.49	&	0.58	&	0.42	&	0.41	&	0.45	&	0.49	\\
	&	Query Rewrite	&	\textbf{0.20}	&	\textbf{0.43}	&	\textbf{0.52}	&	\textbf{0.64}	&	\textbf{0.46}	&	\textbf{0.45}	&	\textbf{0.48}	&	\textbf{0.54}	\\
\bottomrule
\end{tabular}
\caption{\small Retrieval Performance of models on our benchmark using Recall and nDCG metrics}
\label{tab:retrieval-results}
\end{table*}

We use \benchmark\ to evaluate both the retrieval and generation components of RAG systems. Retrieval evaluation is described in this section, while generator evaluation is presented in the next section. Note, that this is not intended to be a comprehensive evaluation of all retrievers and generators. Instead our goal is to demonstrate the challenging nature of \benchmark\ and how it can help evaluate and surface issues in state-of-the-art RAG systems.

\subsection{Experimental Setup}
\label{sec:retrieval-setup}

We use \benchmark{} with the indexed corpora from the creation stage (Section \ref{sec:corpora}) to evaluate lexical (BM25), dense (BGE-base 1.5 \cite{bge_embedding}) and sparse (Elser) retrieval. 
The competitiveness of these models is shown on the MTEB leaderboard.\footnote{\href{https://www.elastic.co/search-labs/blog/elasticsearch-elser-relevance-mteb-comparison}{https://www.elastic.co/search-labs/blog/elasticsearch-elser-relevance-mteb-comparison}} 

We report the commonly used Recall and nDCG metrics ~\cite{thakur2021beir} @1, 3, 5, 10. Since we use Elser for retrieval during data creation, there may be some biases towards Elser.

The retrieval task is performed on the reference passages (i.e., those marked as relevant during the creation process). It is therefore only computed on answerable and partially answerable tasks where reference passages exist.

\subsection{Retrieval strategies}
We experimented with several strategies to query the retriever for relevant passages, including sending the full conversation up to the current user turn, all the user turns without the responses, subsets of the conversation, and only the last user turn. Using the full conversation, or even just a few turns from it consistently under-performed, often causing the retriever to bring back the same passages over and over. The most effective strategy was just using the last user turn. A particular challenge in multi-turn conversations is that a user turn may be non-standalone, employing shortcuts to express intent, or referencing entities or concepts from earlier in the conversation. To alleviate this we used a {\em query rewrite} strategy, also known as contextual query rewriting \cite{Zhou2023,Sun2023}
to rewrite the user turn using an LLM
so that it incorporates all necessary parts from the context into an unambiguous, standalone question (see Appendix \ref{sec:query-rewriting} for implementation details).
An example of query rewrite is shown below:\\

\noindent\fbox{%
    \parbox{\columnwidth-0.25cm}{%
\small
User: Who is the CEO of Apple Inc.? \\
Agent: The CEO of Apple Inc. is TIM COOK. \\
User: its address? \\

\bf{[Rewriting]} What is the address of Apple Inc?
}}\\ 

\subsection{Retrieval Results}
\label{sec:retrieval-results}

We highlight the results in Tables~\ref{tab:retrieval-results} and~\ref{tbl:retrieval-breakdown}. Table~\ref{tab:retrieval-results} shows that the query rewriting strategy consistently outperforms using only the last turn (without rewriting), across all metrics, for all models. 
Elser outperforms BM25 and BGE-base 1.5. 
Table~\ref{tbl:retrieval-breakdown} shows the Elser results with query rewriting broken down by domain, first turn vs later turn, and whether the question is standalone.
Retrieval performance is significantly lower for later turns than for the first turn, and non-standalone questions 
continue to pose a challenge
(though query rewriting helps).
These results highlight two key areas of improvement for retrieval components: 1) multi-turn retrieval, and 2) non-standalone questions.

\begin{table}[ht]
\small
\centering
\begin{tabular}{l||l|l}
\toprule
& {\bf Subset} & {\bf R@5}\\
\midrule
\multirow{ 2}{*}{By Turn} & Turn 1 (102) & 0.89 \\
& > Turn 1 (675) & 0.47 \\
\midrule
By Standalone & Standalone (555) & 0.48 \\
(> Turn 1) & Non-Standalone (120) & 0.42 \\
\midrule
\multirow{ 2}{*}{By Domain} & \clapnq\ (208) & 0.56\\
& FiQA (180) & 0.50 \\
& Govt (201) & 0.56 \\
& Cloud (188) & 0.47 \\
\bottomrule
\end{tabular}
\caption{\small Elser retrieval results with query rewrite on subsets of the data to highlight multi-turn properties. Numbers in parentheses denote size of each subset.}
\label{tbl:retrieval-breakdown}
\end{table}

\section{Generation}
\label{sec:generation}

We next present the generator experiments. We start with the experimental setup, followed by the  results, using automated metrics including LLM judges. Section~\ref{sec:human-eval} will complement this with a human evaluation on a subset of \benchmark. Given a task, we send to the model the following information: the question, preceding turns, N passages, and instructions. We choose N=5 passages because it achieves considerable improvement compared to top 3, while remaining a manageable amount of passages (Section~\ref{sec:retrieval}). For more generation format details, see Appendix~\ref{sec:generation-prompts}.

\subsection{Retrieval Settings}
\label{sec:retrieval-settings}

We evaluate how LLMs perform under three retrieval settings, simulating ideal/noisy retrieval. 

\noindent{\bf Reference  ($\bullet$):} Generation using reference passages or no passages if unanswerable/conversational. No retrieval is performed in this setting; it simulates a perfect retriever.

\noindent{\bf Reference + RAG ($\halfcirc[.4ex]$):} 
Partial retrieval followed by generation, where the reference passages are supplemented by the top retrieved passages (using Elser with rewrite) to yield a total of 5 passages.
We restrict this to the 426 tasks that have at most two reference passages to ensure all passages needed for the reference are included. This can be considered an upper bound, where the retrieval is successful but there are additional noisy passages.

\noindent{\bf Full RAG ($\circ$):} Retrieval using Elser with rewrite followed by generation, where the top N=5 passages are retrieved (the standard RAG setting).

\begin{table*}[!htbp]
\centering
\small
\setlength{\tabcolsep}{5.5pt}
\begin{tabular}{l|ccc||ccc|ccc|ccc}
\toprule
  & \multicolumn{3}{c}{\bf Ans. Acc.} &  \multicolumn{3}{c}{\bf \ragasF}  &  \multicolumn{3}{c}{\bf \radbench}  &  \multicolumn{3}{c}{\bf \agg} \\
& $\bullet$ & $\halfcirc[.4ex]$ & $\circ$ & $\bullet$ & $\halfcirc[.4ex]$ & $\circ$ & $\bullet$ & $\halfcirc[.4ex]$ & $\circ$ & $\bullet$ & $\halfcirc[.4ex]$ & $\circ$ \\
\midrule
Reference  & 0.98 & 0.97 & 0.98 & 0.86 & 0.87 & 0.67 & 0.94 & 0.94 & 0.94 & 0.88 & 0.88 & 0.86 \\
\midrule
Command-R+ (104B)  & 0.86 & \underline{0.86} & \textbf{0.87} & \textbf{0.69} & \textbf{0.71} & \textbf{0.66} & 0.66 & 0.62 & 0.59 & 0.43 & \underline{0.40} & \underline{0.38} \\
GPT-4o  & {\bf 0.89} & \underline{0.86} & \underline{0.86} & \textbf{0.69} & 0.69 & \underline{0.65} & \textbf{0.73} & \underline{0.68} & \textbf{0.66} & \underline{0.46} & \underline{0.40} & \underline{0.38} \\
GPT-4o-mini  & 0.87 & \underline{0.86} & 0.84 & \underline{0.66} & 0.69 & 0.64 & \underline{0.72} & \underline{0.68} & 0.64 & 0.43 & \underline{0.40} & 0.37 \\
Llama 3.1 405B Instruct  & 0.87 & \underline{0.86} & 0.85 & \textbf{0.69} & \underline{0.70} & \underline{0.65} & 0.70 & \underline{0.68} & 0.63 & \textbf{0.47} & \textbf{0.42} & \textbf{0.39} \\
Llama 3.1 70B Instruct  & 0.78 & 0.83 & 0.81 & 0.63 & 0.66 & 0.64 & 0.62 & 0.64 & 0.59 & 0.43 & \textbf{0.42} & \textbf{0.39} \\
Llama 3.1 8B Instruct  & 0.71 & 0.75 & 0.74 & 0.50 & 0.51 & 0.53 & 0.54 & 0.56 & 0.54 & 0.36 & 0.33 & 0.34 \\
Mixtral 8x22B Instruct  & 0.86 & {\bf 0.87} & \underline{0.86} & 0.54 & 0.61 & 0.56 & 0.66 & 0.64 & 0.61 & 0.39 & 0.38 & 0.35 \\
Qwen 2.5 (72B)  & 0.87 & {\bf 0.87} & \textbf{0.87} & 0.65 & \textbf{0.71} & 0.64 & 0.71 & \textbf{0.69} & \underline{0.65} & 0.43 & \underline{0.40} & 0.37 \\
Qwen 2.5 (7B)  & \underline{0.88} & \underline{0.86} & \textbf{0.87} & 0.62 & 0.66 & 0.62 & 0.68 & 0.65 & 0.63 & 0.42 & 0.38 & 0.37 \\
\bottomrule
\end{tabular}
\caption{\small Generation results by retrieval setting: Reference ($\bullet$), Reference+RAG ($\halfcirc[.4ex]$), and RAG ($\circ$), w/ IDK conditioned metrics ($n = 426$). Per column, the best result is in \textbf{bold} and second best is \underline{underlined}.}
\label{tbl:judge_results_gold_by_RAGsetting}
\end{table*}

\subsection{Models}
\label{sec:models}

We evaluate the following auto-regressive models.

\noindent{\bf Llama 3.1 Models}~\cite{dubey2024llama3herdmodels}: The Llama 3.1 family of models are instruction-tuned models that support up to 128K tokens. We evaluate the 8B, 70B and 405B models.

\noindent{\bf Mixtral Mixture-of-Expert Models}~\cite{jiang2024mixtralexperts}:
The instruction fine-tuned Mixtral 8x22B model that supports up to 32K tokens.\footnote{We use the 22B and not the 7B variant of Mixtral, due to the fact that the latter was used during conversation generation.}

\noindent{\bf GPT-4o Models\footnote{\url{https://openai.com/index/gpt-4o-system-card/}}}: We use the GPT-4o and GPT-4o-mini model in our experiments. These support context lengths of up to 128K tokens.

\noindent{\bf Command R+\footnote{\url{https://huggingface.co/CohereForAI/c4ai-command-r-plus}}}: This is a 104B parameter multi-lingual model optimized for RAG and tool use. 

\noindent{\bf Qwen 2.5 models}~\cite{qwen2.5}: We use the instruct versions of the 7B and 72B models. They support a context length of up to 128K tokens.

\begin{figure*}[t]
    \centering
    \begin{subfigure}{.30\textwidth}
    \centering
    \includegraphics[width=.98\textwidth]{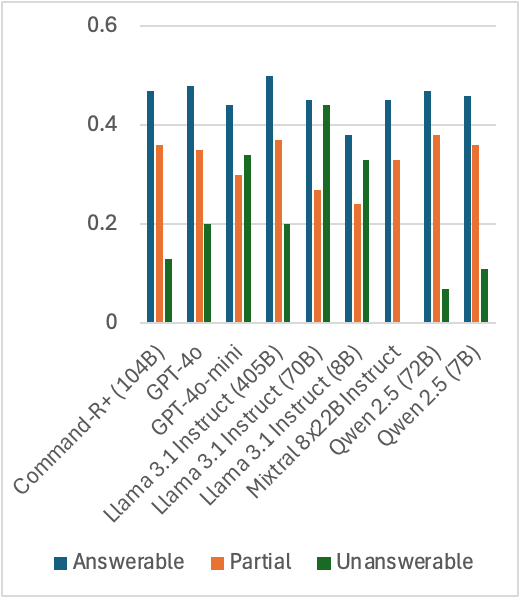}
    \caption{By question answerability}
    \label{fig:generation-breakdown-answerability}
    \end{subfigure}
    \begin{subfigure}{.30\textwidth}
    \centering
    \includegraphics[width=.98\textwidth]{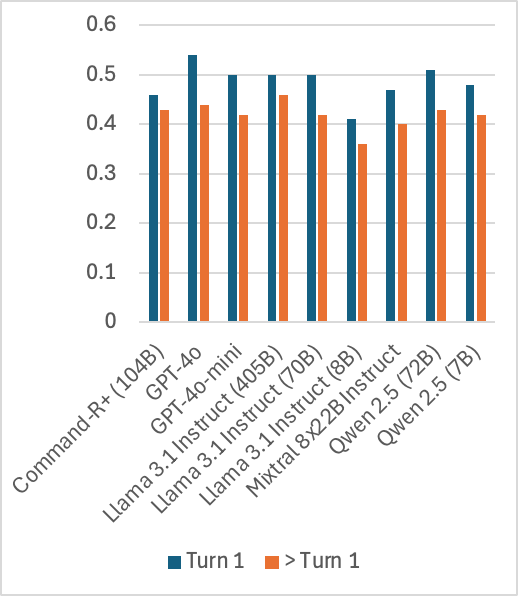}
    \caption{By first vs subsequent turns}
    \label{fig:generation-breakdown-turn}
    \end{subfigure}
    \begin{subfigure}{.37\textwidth}
    \centering
    \includegraphics[width=.98\textwidth]{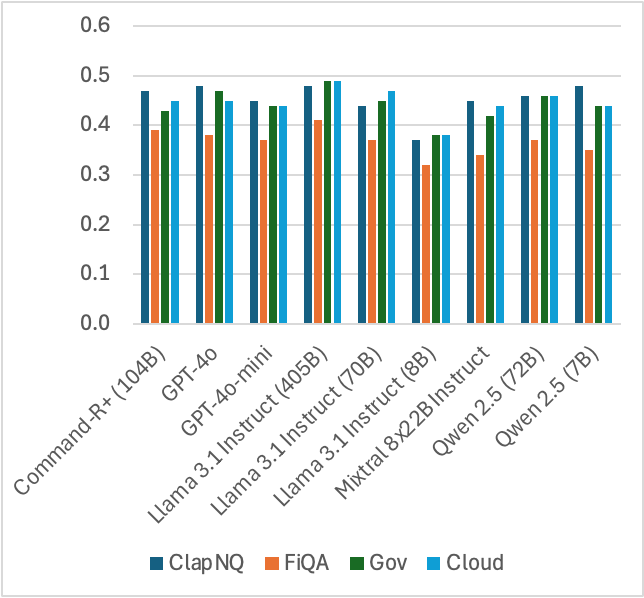}
    \caption{By domain}
    \label{fig:generation-breakdown-domain}
    \end{subfigure}
    \caption{\small Generation results in the Reference ($\bullet$) retrieval setting using a single metric, \agg, on three different dimensions: (a) answerability, (b) turns, and (c) domains}
    \label{fig:generation-breakdown}
\end{figure*}

\subsection{Metrics}
\label{sec:metrics}

We use three metrics, described below, to evaluate the quality of RAG systems and understand if a model response exhibits the desirable \fanc\ properties outlined in Section~\ref{sec:other-characteristics}. We use these metrics to provide insights into the trends of LLMs on multi-turn RAG rather than declare winners. See Section~\ref{sec:evaluation-challenges} for a discussion on how we picked the metrics and the challenges associated with finding good evaluation metrics for multi-turn RAG. Moreover, for implementation details, see Appendix \ref{sec:metrics-details}.

\subsubsection{Aggregate performance metrics}

The first two metrics, \emph{\agg} and \emph{\radbench}, are both reference-based (\textit{RB}) metrics. They both  measure  the aggregate performance of model responses by comparing the model response (\textit{MR}) to the reference answer (\textit{RA}) utilizing different techniques:  

\noindent{\bf \agg} is the harmonic mean of three algorithmic metrics~\cite{adlakha-etal-2024-evaluating}: Bert-Recall (\emph{\bRec}), Bert-K-Precision (\emph{\bkPrec}), and \emph{\rougeL}. The intuition is as follows: \bRec\ is an approximation for completeness as it measures the semantic overlap between \textit{MR} and \textit{RA}, tending to prefer longer answers. \bkPrec\ compares \textit{MR} to the passages, \textit{P}, and is an approximation for faithfulness and completeness. \rougeL\ measures whether phrases from \textit{RA} are in \textit{MR} and is an approximation for appropriateness.

\noindent{\bf \radbench} is an LLM judge inspired by RAD-Bench~\cite{kuo2024radbenchevaluatinglargelanguage}. We adapt RAD-Bench’s approach of comparing \textit{MR} to \textit{RA} but modify the prompt to add \textit{P} and anchor the evaluation on the metrics of faithfulness, appropriateness, and completeness. To minimize model biases and improve evaluation reliability, we use several models as judges and use the median as the final score.

\subsubsection{Faithfulness metric}

We also use a metric specifically for faithfulness, which is 
important for RAG applications
and a challenge for LLMs (see Section ~\ref{sec:human-eval}).
\noindent{\bf \ragasF}, the Faithfulness LLM judge from RAGAS \cite{es-etal-2024-ragas}, appears to be a good judge for faithfulness (see Section~\ref{sec:evaluation-challenges}).
In contrast to the other two metrics, this is a reference-less (RL) metric, as it does not rely on the reference answer. 

\subsubsection{Conditioning metrics on answerability}
\label{sec:answerability-idk-conditioning}

Prior to computing the metrics we employ an IDK ("I Don't Know") judge to detect whether the response has a full or partial answer. It is important to first determine whether a response is IDK because intuitively, words used to indicate not knowing the answer may not match the context; this is also reflected in the metrics, which were not designed to measure IDK correctly.
Our IDK judge achieves an accuracy of over 97\% (see Appendix~\ref{sec:idk-judge}).
We condition the metric score, 
 $\phi$ $\in \{$\agg, \radbench, \ragasF$\}$, on answerability value $A$ and IDK value $IDK$ as follows:
 
\begin{table}[h]
\centering
\small
\begin{tabular}{l | c c}
    \toprule
     & $IDK$ = no, partial & $IDK=$ yes \\
    \midrule
    $A=$ yes, partial & $\phi$ & 0 \\
    $A=$ no & 0 & 1 \\
    \bottomrule
\end{tabular}
\end{table}

We define \emph{answerability accuracy} (\emph{Ans. Acc.}) of a model as the accuracy of $IDK$ correctly predicting $A$. Finally, conversational questions also require special handling for evaluation. Since only 10 tasks fall in this category, we exclude them from the experiments, leaving their study as future work.

\subsection{Generation: Evaluation Results}

\label{sec:model-behavior-challenges}

The overall results across the different retrieval settings are shown in Table~\ref{tbl:judge_results_gold_by_RAGsetting}.\footnote{These results are limited to the 426 Reference + RAG ($\halfcirc[.4ex]$) tasks to be consistent across retrieval settings. Appendix~\ref{sec:generation-tables} shows that the trends persist throughout the benchmark.}
Within model families, the larger model does as well or better than its smaller counterparts across metrics. In general, GPT-4o and Llama 3.1 405B Instruct perform the best across metrics and settings. 
All models score significantly lower than the reference answer,
indicating there is still room for improvement in \benchmark\ for all LLMs.
Comparing performance across retrieval settings, we see that
the results degrade as the  setting gets more challenging: $\bullet > \halfcirc[.4ex] > \circ$, indicating the noise has an impact on generation. Interestingly, Qwen 2.5 72B and Command-R+ are more competitive and closer to GPT-4o and Llama models in noisy settings. 

Focusing next on the Reference ($\bullet$) retrieval setting, we further explore how LLMs perform on the different dimensions of \benchmark. Figure~\ref{fig:generation-breakdown} shows the results. For space reasons we only report the \agg metric. We also leave out the breakdown by question types and multi-turn types, where we did not find interesting patterns. The additional metrics and breakdowns can be found in Appendix \ref{sec:generation-tables}. 

\noindent{\bf Answerable vs unanswerable questions.} 
As shown in Figure~\ref{fig:generation-breakdown-answerability}, model performance generally drops when questions become partially answerable or unanswerable. In particular, models experience a dramatic drop in performance on unanswerables, struggling to declare they do not know the answer when the answer is not included in the input passages. Interestingly, performance on unanswerables differs widely between model families. While GPT-4o and Llama 405B score low for unanswerables, they still perform much better than models in other families. It is interesting to note that Llama 70B and 8B perform better on the unanswerables, because they say ``I don't know'' too often. This can also be seen by their low answerability accuracy in Table~\ref{tbl:judge_results_gold_by_RAGsetting}.

\noindent{\bf First turn vs subsequent turns.} As shown in Figure~\ref{fig:generation-breakdown-turn}, in almost all cases the models perform better on first turn vs subsequent turn questions. This proves our conjecture that answering questions in a multi-turn setting is more challenging, as a model has to interpret a question in the context of the preceding conversation, which was one of the main motivations for this work.

\noindent{\bf Domains.} 
Figure~\ref{fig:generation-breakdown-domain} shows that model performance is similar across domains except for FiQA where the results tend to be lower. We suspect this is due to the nature of the corpus; it contains posts from a financial discussion forum, which are typically short, very informal in style, and often subjective.

\section{Human Evaluation}
\label{sec:human-eval}

We also performed a human evaluation on a subset of the benchmark. This serves several purposes: 1) Verifying that the reference responses are of high quality, 2) correlating our metrics with human judgment, and 3) analyzing frontier models. 

We perform the human evaluation on two frontier models: GPT-4o and Llama 3.1 405B Instruct and compare them to the reference answers. We select 5 conversations per domain for a total of 159 evaluation tasks. We ask annotators to measure the quality across the desirable response properties, FANC (Section~\ref{sec:other-characteristics}),
on a scale of 1 (Low) - 4 (High). We also ask them to perform a pair-wise comparison of the models, from which we calculate the \emph{Win-Rate (WR)}. Ties are allowed but we ask them to do so sparingly. 

\begin{table}[t]
\centering
\setlength{\tabcolsep}{2.2pt}
\small
\begin{tabular}{l|r|rrrrr|rrr}
\toprule
 & \multicolumn{6}{|c}{Human Evaluation} & \multicolumn{3}{|c}{Metrics} \\
& WR & F & A & N & C &  All & \scriptsize{\ragasF} & \scriptsize{\radbench} & \scriptsize{\agg} \\
\midrule
Ref. & \textbf{59.3} & \textbf{3.8} & 3.8 & \textbf{4.0} & \textbf{3.9} &  \textbf{3.8} & \textbf{0.86} & \textbf{0.97} & \textbf{0.87} \\
GPT-4o & 47.8 & 3.5 & 3.8 & \textbf{4.0} & 3.7 & 3.6 & 0.80 & 0.80 & 0.46 \\
Llama3.1 & 47.4 & 3.5 & \textbf{3.9} & 3.9 & 3.7 & 3.6 & 0.78 & 0.78 & 0.49 \\
\bottomrule
\end{tabular}
\caption{\small Results on the human evaluation for win-rate (WR), \textbf{F}aithfulness, \textbf{A}ppropriateness, \textbf{N}aturalness, \textbf{C}ompleteness and All (harmonic mean of FANC) compared to the IDK conditioned metrics. All results are reported on the Answerable subset.}
\label{tbl:judge_results_gold_human_llmjudge2}
\end{table}

The annotators completed the task using the Appen platform.\footnote{\url{https://www.appen.com/}} Each evaluation task comprised the conversation thus far up to the last user question, the relevant passages, and the corresponding model responses (anonymized and in random order). Each task was done by three skilled annotators. The annotator agreement was very high: \{F: 89.6, A: 92.1, N: 95.6, C: 90.3, WR: 87.4\} (see Appendix \ref{sec:annotator-agreement}).

Table~\ref{tbl:judge_results_gold_human_llmjudge2} shows the evaluation results using median to consolidate the scores of the three annotators on the answerable/partial subset (152 tasks) of the human evaluation. The reference answer is exceedingly preferred by annotators over the model responses, as evidenced by the win-rate. It also exhibits the highest score for most individual properties. This highlights the quality of the human-generated reference answers and shows that even frontier LLMs still have room for improvement. The results show that LLMs do a good job of providing natural and appropriate answers; 
but they still struggle with faithfulness and completeness, where they receive lower scores. Both frontier models are equally preferred for answerable questions. 

We also explore the few unanswerable questions, and observe that Llama 3.1 405B Instruct is the least preferred by win-rate. Llama still provides answers (hallucinations) to unanswerable questions.

\section{Automatic Evaluation}
\label{sec:evaluation-challenges}

\begin{table}[t]
\small
\centering
\begin{tabular}{l|rr|rrr}
\toprule
 & \agg & \radbench & \ragasF & \ragasR & \mtb \\
\midrule
WR & 0.24  & 0.33 & 0.01 & -0.03 & -0.17 \\
\bottomrule
\end{tabular}
\caption{\small Weighted Win-Rate Spearman correlation with reference-based (RB) and reference-less (RL) metrics}
\label{tab:winrate-correlation}
\end{table}

\begin{figure}
    \centering
    \includegraphics[width=0.75\linewidth]{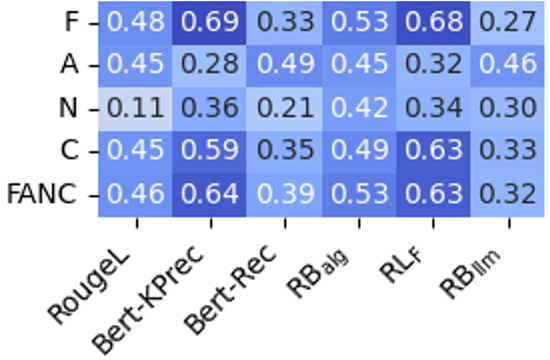}
  \caption{\small Weighted Spearman correlation of human evaluation with the automated metrics on the answerable subset for the GPT-4o and Llama 3.1 405B Inst. models.}
    \label{fig:correlation}
\end{figure}

Human evaluation is not feasible as a long term solution for evaluating models as it does not scale easily. We explore reference-less and reference-based automated evaluation via algorithmic metrics and LLM judges. Table~\ref{tab:winrate-correlation}
shows the correlation of these metrics with the human evaluation's win-rate and Figure~\ref{fig:correlation} shows the correlation of metrics that correlate positively with win-rate for the individual properties of FANC. We employed a weighted correlation on human judgements to correct for data imbalance. We exclude the reference responses from Figure~\ref{fig:correlation} because by design the ranking of the reference is 1 for the Reference-Based metrics. These findings drive our decision to report \agg , \radbench, and \ragasF\ as our main metrics in Section~\ref{sec:generation}.

\noindent{\textbf{Reference-Based Metrics}.} We consider an adapted version of RAD-Bench \cite{kuo2024radbenchevaluatinglargelanguage}, denoted as \radbench, as well as the harmonic mean of algorithmic metrics, denoted as \agg, as described in Section~\ref{sec:metrics}. As shown in Table~\ref{tab:winrate-correlation}, these metrics correlate reasonably well with win-rate; in particular \radbench. Figure~\ref{fig:correlation} shows that they also tend to correlate well with our desired properties; in particular \agg.

\noindent{\textbf{Reference-Less Metrics}.} We investigate popular metrics in literature: the Faithfulness (\ragasF) and Answer Relevance (\ragasR) metrics from RAGAS~\cite{es-etal-2024-ragas}, an adapted version of MT-Bench (\mtb\, as described in Appendix \ref{sec:mtbench-judge}). \cite{zheng-mt-bench-2024}, as well as \bkPrec.
RAGAS Answer Relevance (\ragasR), evaluates the model response by asking the LLM to determine the question associated with the response. This question is then compared with the actual question. This is not suitable in a multi-turn setup where the question is often non-standalone. Similarly, \mtb, while adapted to fit multi-turn conversations, has a very low correlation with win-rate, favoring machine-generated text. In contrast, we find that the two metrics focusing on faithfulness, \bkPrec\ and  RAGAS Faithfulness (\ragasF), correlate well with the human faithfulness scores and are therefore useful in settings where no reference is available, although it is clear that evaluating only the faithfulness aspect of a response is incomplete.

\section{Synthetic Conversations}
\label{sec:synthetic}

\begin{table}[t]
\centering
\small
\setlength{\tabcolsep}{2.1pt}
\begin{tabular}{l r r}
    \toprule
     & {\bf \benchmark} & {\bf \benchmarksynth}\\
    \midrule
    Avg \# turns per conversation  & 7.7 & 5.9 \\
    Avg \# unique passages per conv. & 16.9 & 4.6 \\
    Avg \# edited responses per conv. & 7.3 & - \\
    \midrule
    Avg \# of question words & 8.6 & 13.7 \\
    Avg \# of response words * & 97.2 & 72.4 \\
    Avg \# of relevant passages * & 2.8 & 4.1 \\
    \midrule
    Total \# of Conversations & 110 & 200\\
    Total \# of Tasks & 842 & 1,181\\
    \bottomrule
\end{tabular}
\caption{\small Comparing human-generated \benchmark\ and synthetically-generated \benchmarksynth. (Note: Properties with * computed on answerable+partial subset)}
\label{tbl:overall-stats}
\end{table}

Manually creating data is an expensive and time-consuming process that does not scale well. Automating this process has become popular via synthetic data generation~\cite{soudani2024surveyrecentadvancesconversational} and can serve as useful evaluation. To explore this direction, we construct a companion benchmark, \benchmarksynth\ of synthetically-generated conversations. We extend the recently proposed framework of \cite{leegunasekara2024} to automatically generate multi-turn conversations (see Appendix \ref{sec:synthetic-implementation-details} for details). To ensure that the two benchmarks are comparable, we utilize the same corpora, question types, multi-turn patterns, and answerability types (Section~\ref{sec:dimensions}). 

In Table \ref{tbl:overall-stats}, we see that synthetic conversations are typically shorter (averaging 5.9 vs 7.7 turns) and exhibit a lower passage diversity (4.6 vs 16.9 unique passages per conversation). Question and response lengths also differ with synthetic data having longer (potentially more detailed) questions but shorter answers. We also found that mimicking several important characteristics posed challenges: our attempts to synthetically generate unanswerable questions were not very successful as the model would often create questions with at least a partial answer. Moreover, increasing the number of turns tended to lead to repetitive user questions and a higher likelihood of hallucinated agent responses.

Since the aggregate metrics (\agg\ and \radbench) rely on a reference answer, which does not exist for synthetic data, we employ the reference-less \ragasF\ and \bkPrec\ to evaluate faithfulness on \benchmarksynth. Using either metric, we see that models across the board receive a higher faithfulness score on the synthetic than on the human-generated data (see Appendix \ref{sec:faithfulness-synthetic}). There are multiple potential explanations for this,
ranging from potential idiosyncrasies of synthetic data generation approaches, to the reliability of the presence of desired characteristics, to the quality and biases of automatic evaluation metrics. More work is needed to compare human-generated and synthetic data and we hope that our companion synthetic benchmark serves as a valuable asset towards that goal.
\section{Conclusions and Future Work}

We present \benchmark, a comprehensive and diverse benchmark of 110 multi-turn human-generated conversations averaging 7.7 turns for a total of 842 tasks.~\footnote{\benchmark\ is publicly available at \url{https://github.com/ibm/mt-rag-benchmark}.} These tasks are used to test the full RAG pipeline. \benchmark\ is the first end-to-end human-generated multi-turn RAG benchmark that reflects real-world properties of multi-turn conversations.
Our experimental results, employing both automated metrics and a human evaluation, highlight the quality of our benchmark and outline several trends and challenges related to multi-turn RAG systems that state-of-the-art retrievers and LLMs face during retrieval and generation. Our findings encourage future research on improving retrieval and generation performance, especially in longer multi-turn conversations, unanswerable questions, and non-standalone user questions. In addition, the ability to scale indicates a clear need for i) more accurate reference-less automatic evaluation metrics, which align more closely with human judgement and can better differentiate model performance;  and ii) synthetic data to obtain more conversations. We are also motivated to extend \benchmark\ in the future to include adversarial turns, additional domains and multilingual conversations.

\clearpage
\bibliography{custom}
\bibliographystyle{acl_natbib}

\clearpage

\appendix
\section{Dimension definitions}
\label{sec:diff-types-def}

This section provides a detailed list of the \benchmark's dimensions, introduced in Section \ref{sec:dimensions}. In particular, Tables \ref{tbl:question-types-def}, \ref{tbl:multi-turn-types-def}, and \ref{tbl:answerability-types-def} show the definitions of the question types, multi-turn types, and answerability types included in the benchmark. 

\begin{table*}[p]
\centering
\scriptsize
\begin{tabular}{p{2.2cm} p{7cm} p{6cm}}
    \toprule
    {\bf Question type} & {\bf Definition} & {\bf Example questions}\\
    \midrule
    Comparative & Asking for comparison. This can be comparison (a) of multiple entities/concepts, (b) of characteristics of a single entity, or (c) comparison with decision. & (a) “What's the difference between effective and marginal tax”, (b) “pros and cons of credit cards”, (c) “is X better than Y”\\
    \midrule
    Composite & Comprises several questions. They may be related or dependent. & “Am I eligible for a driver’s license and how do I apply?”\\
    \midrule
    Explanation & Explain the reason behind something. & “Why do I have to do/have X…?”\\
    \midrule
    Factoid & Asking for a specific piece of information, such as a date, quantity, name, yes/no answer, or other singular fact. It can be answered directly and concisely, and does not require an explanation, opinion, interpretation, or subjective judgment to answer. & “Am I eligible for a driver’s license?” , “What is the link to the application portal?”\\
    \midrule
    How-To & Instructions describing how to perform a task. & “How should I do X …” , “How do I apply for social security disability benefits?” , “I need a license. What do I need to do?”\\ 
    \midrule
    Keyword & Asking using keywords (not full sentence/phrase). This may be ambiguous. & “vacation days”, “ios 17 upgrade”\\
    \midrule
    Non-question & Not asking a question but instead answering a question or providing information asked by the model. & “I am in Sacramento”, in response to the model saying: "The procedure to file a restraining order depends on the type of restraining order and the court where you want to file it in."\\    
    \midrule
    Opinion & Asking model’s opinion on something. The question could also be phrased as leading. & “Don’t you think iphone is better than samsung?”, “Which color car is the best?”\\
    \midrule
    Summarization & Asking to summarize a process or a policy. & “What’s the policy on vacation days?”\\
    \midrule
    Troubleshooting & Finding solutions to issues, problems, challenges. & “I have the error X… what should I do?”\\
    \bottomrule
\end{tabular}
\caption{\small Definitions of question types}
\label{tbl:question-types-def}
\end{table*}

\begin{table*}[p]
\centering
\scriptsize
\begin{tabular}{p{1.6cm} p{4.3cm} p{8.7cm}}
    \toprule
    {\bf Multi-turn type} & {\bf Definition} & {\bf Example}\\
    \midrule
    Follow-up & Ask a question that requests more information or related information to continue the conversation. &
    $\sbullet$ User: Can you tell me about the responsibilities of SSA?\newline
    $\sbullet$ Agent: Social Security Administration (SSA) assigns Social Security numbers and runs Social Security retirement and disability insurance programs.
    \newline
    $\sbullet$ {\em User [Follow-up]: I have forgotten my social security number. Should I contact SSA regarding that?}\\
    \midrule
    Clarification & Clarify user’s intent or model’s answer:
    \newline\newline
    (a) Write a statement clarifying the user’s intent (typically used when the agent misinterpreted one of the prior questions).
    \newline\newline
    (b) Ask a question to clarify the model’s previous answer. Clarifications are typically asked when something is unclear or hard to understand. &
    (a) Clarification of user’s intent: 
    \newline
    $\sbullet$ User: graphql
    \newline
    $\sbullet$ Agent: GraphQL is an open-source data query and manipulation language for APIs and a query runtime engine.
    \newline
    $\sbullet$ {\em User [Clarification]: No, I meant, how do I set it up.}
    \newline\newline
    (b) Clarification of model’s answer:
    \newline
    $\sbullet$ User: Can you tell me about the responsibilities of SSA?
    \newline
    $\sbullet$ Agent: Social Security Administration (SSA) assigns Social Security numbers and runs Social Security retirement and disability insurance programs.
    \newline
    $\sbullet$ {\em User [Clarification]: Can you explain what you meant by “Social Security retirement and disability insurance programs"?}\\
    \bottomrule
\end{tabular}
\caption{\small Definitions of multi-turn types}
\label{tbl:multi-turn-types-def}
\end{table*}

\begin{table*}[thp]
\centering
\scriptsize
\begin{tabular}{p{3cm} p{12cm}}
    \toprule
    {\bf Answerability type} & {\bf Definition}\\
    \midrule
    Answerable & The question can be fully answered from the passages.\\
    \midrule
    Partially answerable & Only part of the question can be answered from the passages.\\
    \midrule
    Unanswerable & The question cannot be answered neither fully nor partially from the passages.\\
    \midrule
    Conversational & The user turn does not contain a question but is a conversational statement (e.g., "Hello", "Hi, I had a question", "Cool", "That's interesting", "That was all", "Thank you")\\
    \bottomrule
\end{tabular}
\caption{\small Definitions of answerability types}
\label{tbl:answerability-types-def}
\end{table*}

\section{Benchmark creation details}

\subsection{Creating the \govt\ and \cloud\ corpora}
\label{sec:corpora-details}

The \cloud\ corpus was created by crawling the  public technical documentation of a major cloud provider, thus containing inter-connected pages covering various aspects of each cloud offering.

Similarly, the \govt\ corpus was created starting from 100 seed web pages from the .gov and .mil domains, selected to be suitable to write conversations on (e.g., excluding very short pages or navigation pages) and diverse in terms of topics (e.g., covering parks, NASA, Veteran Affairs, city web-sites, DMV, etc.). For each seed page, we crawled web pages appearing in its neighborhood and thus expected to cover related topics. For each neighborhood the top 150 pages were selected based on their page rank score and additionally filtered to remove duplicates and other low-quality pages, leading to the final set of 7,661 pages/documents.

\subsection{Creating conversations}

The following instructions were given to the annotators for creating conversations:

\noindent\fbox{%
    \parbox{\columnwidth-0.25cm}{%
\scriptsize
\begin{itemize}[leftmargin=.3cm]
    \item Given a seed question, continue a conversation between you and the agent.  Continue the conversation to generate a total of 6-9 turns (turn = question + response). 
    \item Aim to have around 25\% unanswerable questions. There should be no more than 2 unanswerable questions in one conversation. Most conversations should be completely answerable.
    \item We encourage diversity of questions across types.
    \item  The questions in your conversation should be connected, but diverse enough to include different passages. All questions shouldn't have the same relevant passages.
    \item The agent is a bot and the answers may be wrong. In addition to creating questions, you will also edit the answers if they are wrong. Answer correctness is based on the passages provided with the answer. Good conversations will have questions that require edits to repair the conversation.
\end{itemize}
}}\\

\subsection{Reviewing conversations}
\label{sec:conv-review}

As described in Section \ref{sec:human-gen-conversations}, the conversations initially generated by annotators went through a round of review before being finalized. During the review process, 16 out of the 126 conversations were rejected due to lack of turn-wise and topic-level coherency as well as passage diversity, leading to 110 accepted conversations. Additionally, during review, 264 responses (31.3\%) were repaired, leading to a slight increase on the number of repaired responses in the final benchmark (from 778 repaired responses before review to 799 responses after review, respectively). The average \textit{edit score} on these responses also increased from 59.5 to 69.2.

\section{Retrieval experiment details}

\subsection{Query rewriting}
\label{sec:query-rewriting}

\begin{figure}[h]
    \centering
    \includegraphics[width=\linewidth]{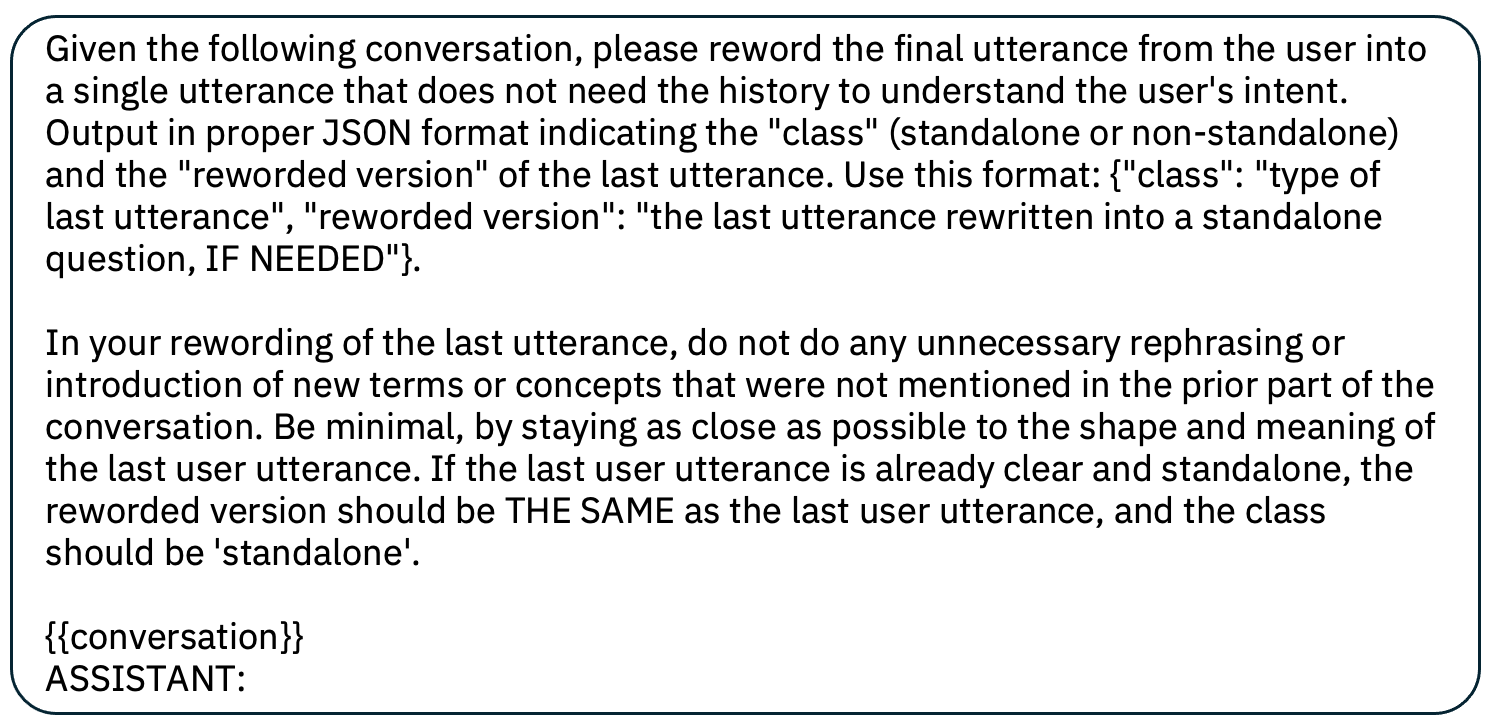}
    \caption{\small Query rewrite prompt}
    \label{fig:qw-prompt}
\end{figure}

\noindent
For the retrieval experiments, we implemented query rewriting by sending the prompt of Figure~\ref{fig:qw-prompt} to Mixtral 8x7B Instruct. This is only a reference implementation, and better ones could be obtained with larger models (e.g., Mixtral 8x22B Instruct, Llama 3.1 405B, etc.), few-shot prompts with in-context-learning examples, or via fine-tuning. However, this  implementation is sufficient to make the point in Section~\ref{sec:retrieval} that query rewriting is both necessary and effective in mitigating the challenges of non-standalone questions in multi-turn data. 

\section{Generation experiment details}

\subsection{Model choices}

To pick the models used in the generation experiments, we selected a suite of state-of-the-art models of varying sizes. It is not an exhaustive list, but shows the common trends in frontier models. For OpenAI models, we also experimented with the latest models in the o1~\footnote{\url{https://platform.openai.com/docs/models\#o1}} family. However, initial experiments showed that o1 did not perform as well as GPT-4o. We suspect that this is due to o1's focus on reasoning and multimodal rather than on RAG.

\subsection{Model invocation}
\label{sec:generation-prompts}

To evaluate LLMs on \benchmark, for each task we sent to the LLM the following information: question, preceding turns, passages, and instruction. This was composed into the  prompt below, adapted to different models using HuggingFace's ChatTemplate\footnote{\url{https://huggingface.co/docs/transformers/main/en/chat_templating}}. In all cases the prompt fit within the model's context length without need for truncation:

\noindent\fbox{%
    \parbox{\columnwidth-0.25cm}{%
\scriptsize
Given one or more documents and a user query, generate a response to the query using less than 150 words that is grounded in the provided documents. If no answer can be found in the documents, say, ``I do not have specific information'' 

\textbf{PASSAGE 1}

...

\textbf{PASSAGE M}

\textbf{User turn 1}

\textbf{Agent Turn 1}

... 

\textbf{User Turn N}
}}\\

\subsection{Example output}

Figure \ref{fig:example-task-with-responses} shows an example of an \benchmark\ task using the Reference ($\bullet$) retrieval setting together with the Reference response as well as responses from several LLMs.

\begin{table*}
\scriptsize
	\begin{tabular}{p{15.5cm}}
	\toprule
	\multicolumn{1}{|c|}{\textbf{Conversation}} \\ 
	\toprule 
	\textbf{User}: where does doctor strange get his powers from \\ 
	\textbf{Agent}: Doctor Strange's powers come from mystical entities such as Agamotto, Cyttorak, Ikonn, Oshtur, Raggadorr, and Watoomb, who lend their energies for spells. He also wields mystical artifacts including the Cloak of Levitation, the Eye of Agamotto, the Book of the Vishanti, and the Orb of Agamotto which give him additional powers. \\ 
	\textbf{User}: did he always have powers? what was he before? \\ 
	\textbf{Agent}: No, Doctor Strange did not always have powers. He was a brilliant but egotistical neurosurgeon. He was in a car accident that ended his surgical career. He searched the globe for a way to repair his hands and encountered the Ancient One, eventually becoming one of her students. \\ 
	\textbf{User}: how many films does he appear in \\ 
	\toprule 
	\multicolumn{1}{|c|}{\textbf{Passages}} \\ 
	\toprule 
	\textbf{Passage 1} \\ 
	\midrule 
The character appears in the Marvel Cinematic Universe. He is first referenced as merely `` Stephen Strange '' in dialogue by Agent Sitwell in the 2014 film Captain America: The Winter Soldier , and first appears in the 2016 film Doctor Strange , in which he is portrayed by Benedict Cumberbatch . He reprises the role in the films Thor: Ragnarok and Avengers: Infinity War , and will appear again in the untitled sequel to Avengers: Infinity War. Doctor Strange appears in the 2016 animated film Hulk: Where Monsters Dwell, voiced by Liam O'Brien. \\ 
	\toprule 
	\textbf{Passage 2} \\ 
	\midrule 
In 2008 , Doctor Strange was ranked 83rd in Wizard 's `` 200 Greatest Comic Book Characters of All Time '' list , and in 2012 was ranked 33rd in IGN 's list of `` The Top 50 Avengers '' . He was also ranked 38th on IGN 's list of `` Top 100 Comic Book Heroes '' . The character was first portrayed in live - action by Peter Hooten in the 1978 television film Dr. Strange . Benedict Cumberbatch stars as the character in the Marvel Cinematic Universe , first appearing in the role in the 2016 film Doctor Strange . He reprised the role in the 2017 film Thor : Ragnarok , the 2018 film Avengers : Infinity War , and will return in the fourth Avengers film in 2019 . \\ 
	\toprule 
	\textbf{Passage 3} \\ 
	\midrule 
Strange appeared as a supporting character for the bulk of the 2000s . He appeared regularly in The Amazing Spider - Man under J. Michael Straczynski , before being cast into a time loop by Baron Mordo . He later appeared on and off in The New Avengers , where he was stated as being part of the secret group known as the Illuminati to deal with future threats to Earth . Ultimately Strange joined the team and allowed them to use his home as a base after the events of Civil War , which he sat out . Doctor Strange was critical of the federal Superhuman Registration Act and aided the anti-registration Avengers team led by Luke Cage . \\ 
	\toprule 
	\multicolumn{1}{|c|}{\textbf{Responses}} \\ 
	\toprule 
	\textbf{Reference} \\ 
	\midrule 
	I do not have an exact count of films in which Doctor Strange appears, but here are some: Doctor Strange, Thor: Ragnarok, Avengers: Infinity War, and The Amazing Spider-Man. Doctor Strange, voiced by Liam O'Brien, appears in the 2016 animated film Hulk: Where Monsters Dwell and will appear again in the untitled sequel to Avengers: Infinity War, which will return in the fourth Avengers film in 2019. \\ 
	\toprule 
	\textbf{Llama 3.1 405B} \\ 
	\midrule 
	According to the text, Doctor Strange appears in the following films in the Marvel Cinematic Universe: 

1. Doctor Strange (2016)
2. Thor: Ragnarok (2017)
3. Avengers: Infinity War (2018)
4. The untitled sequel to Avengers: Infinity War ( Avengers: Endgame, released in 2019, although not explicitly mentioned in the text) Additionally, he appears in the 2016 animated film Hulk: Where Monsters Dwell and the 1978 television film Dr. Strange, played by Peter Hooten. \\ 
	\toprule 
	\textbf{GPT-4.0} \\ 
	\midrule 
	Doctor Strange, portrayed by Benedict Cumberbatch, appears in the following Marvel Cinematic Universe films:

1. **Doctor Strange** (2016)
2. **Thor: Ragnarok** (2017)
3. **Avengers: Infinity War** (2018)
4. **Avengers: Endgame** (2019, referenced as the fourth Avengers film)
5. **Doctor Strange in the Multiverse of Madness** (subsequent appearances expected, although not specified in the provided documents) He also appears in the 2016 animated film *Hulk: Where Monsters Dwell*, voiced by Liam O'Brien. \\ 
	\toprule 
	\textbf{Command-R+ (104b)} \\ 
	\midrule 
	According to the information provided, Benedict Cumberbatch first portrayed the character of Doctor Strange in the Marvel Cinematic Universe in 2016. Cumberbatch reprised the role in two other films, "Thor: Ragnarok" and "Avengers: Infinity War", and will appear again in the untitled sequel to "Avengers: Infinity War". Doctor Strange also appears in the 2016 animated film "Hulk: Where Monsters Dwell", voiced by Liam O'Brien. Doctor Strange was first portrayed in live action by Peter Hooten in a 1978 television film. \\ 
	\toprule 
	\textbf{Qwen 2.5 (72b)} \\ 
	\midrule 
	Doctor Strange, portrayed by Benedict Cumberbatch, appears in at least four films within the Marvel Cinematic Universe: "Captain America: The Winter Soldier" (referenced), "Doctor Strange" (2016), "Thor: Ragnarok" (2017), and "Avengers: Infinity War" (2018). He is also set to appear in the 2019 sequel to "Avengers: Infinity War". Additionally, he appears in the 2016 animated film "Hulk: Where Monsters Dwell", voiced by Liam O'Brien. \\ 
	\bottomrule 
	\end{tabular}
 \caption{\small Example of an \benchmark\ task using the Reference ($\bullet$) retrieval setting (this task is part of the conversation shown in Figure~\ref{fig:conversation_example}) with the Reference response as well as responses from several LLMs. It is clear that several models hallucinate.}
 \label{fig:example-task-with-responses}
\end{table*}

\section{Annotator agreement}
\label{sec:annotator-agreement}

We next describe the computation of the annotator agreement reported in the human evaluation (Section \ref{sec:human-eval}). For each property, the annotator agreement  was computed as the percentage of tasks on which the 3 annotators had either an \emph{absolute agreement} (i.e., they all agreed on the score) or \emph{high agreement} (i.e., two of the annotators agreed on the score and the third annotator gave a score that was a single point away from the score given by the majority).  

\section{Metrics details}
\label{sec:metrics-details}

\subsection{\radbench\ implementation}
\label{sec:rad-implementation}

To adapt the RAD-Bench judge \cite{kuo2024radbenchevaluatinglargelanguage} to our RAG setting, we make two modifications: (1) We include the retrieved passages as additional input to the evaluation, and (2) we anchor the evaluation on our desired properties of faithfulness, appropriateness, and completeness.
Figure~\ref{fig:radbench-prompt} shows the final prompt that we used to implement \radbench. Finally, in order to minimize model biases and improve evaluation reliability, we use four models as judges: GPT-4o-mini (2024-07-18), Qwen 2.5 (72B), Mixtral 8x22B Instruct, and Llama 3.1 405B Instruct, taking the median as the final score.

\begin{figure}[!ht]
\noindent\fbox{%
    \parbox{\columnwidth-0.25cm}{%
\scriptsize
[Instruction]

Please act as an impartial judge and evaluate the quality of the response provided by an AI assistant to the user
question given the provided document and a reference answer.
\par\medskip
Your evaluation should assess the faithfulness, appropriateness, and completeness. Your evaluation should focus on
the assistant's answer to the question of the current turn. You will be given the assistant's answer and a
reference answer. You will also be given the user questions and assistant's answers of the previous turns of the
conversation. You should consider how well the assistant's answer captures the key information, knowledge points
mentioned in the reference answer, and how it respects or builds upon the focus and knowledge points from the
previous turns.
\par\medskip
[Faithfulness]: You are given the full conversation, the question of the current turn, the assistant's answer, and
documents. You should evaluate how faithful is the assistant's answer to the information in the document and
previous conversation.

[Appropriateness]: You should evaluate if the assistant's answer is relevant to the question of the current turn
and if it addresses all the issues raised by the question without adding extra information.

[Completeness]: You should evaluate whether the assistant's answer is complete with information from the documents.
\par\medskip
Begin your evaluation by comparing the assistant's answer against the reference answer in this turn. Be as
objective as possible, and provide a detailed justification for your rating. After providing your explanation, you
must rate the response on a scale of 1 to 10, strictly following this format: "Rating: [[rating]]", for example:
"Rating: [[5]]".
\par\medskip
[The Start of Previous Conversation]

\{previous\_conversation\}

[The End of Previous Conversation]
\par\medskip
[The Start of Current Turn Question]

\{current\_question\}

[The End of Current Turn Question]
\par\medskip
[The Start of Reference Answer]

\{reference\_answer\}

[The End of Reference Answer]
\par\medskip
[The Start of Assistant's Answer]

\{response\}

[The End of Assistant's Answer]
\par\medskip
[The Start of Document]

\{passages\}

[The End of Document]
}}
\caption{\small Prompt used for the \radbench\ judge}
    \label{fig:radbench-prompt}
\end{figure}

\subsection{\mtb\ implementation}
\label{sec:mtbench-judge}

The MT-Bench judge was originally designed for evaluating conversations containing exactly two turns (i.e., user question/agent response pairs), with a focus on properties such as helpfulness, depth, and creativity~\cite{zheng-mt-bench-2024}. 
To use it for assessing general multi-turn conversations with relevant passages, we make the following changes: (1) We focus the evaluation on the last user turn of a multi-turn conversation, (2) we include retrieved passages as additional input, and (3) we modify the instruction to include desired properties, such as faithfulness and completeness. Figure~\ref{fig:mtbench-prompt} shows the final prompt that we used to implement \mtb.

\begin{figure}[ht]
\noindent\fbox{%
    \parbox{\columnwidth-0.25cm}{%
\scriptsize
[System]

Please act as an impartial judge and evaluate the quality of the response provided by an AI assistant to the user
question given the provided document. Your evaluation should consider factors such as the faithfulness,
appropriateness, conciseness, relevance, accuracy, completeness, and correctness of their responses. You evaluation
should focus on the assistant's answer to the final user question in the conversation. Begin your evaluation by
providing a short explanation. Be as objective as possible. After providing your explanation, you must rate the
response on a scale of 1 to 10 by strictly following this format: [[rating]], for example: Rating: [[5]].
\par\medskip
<|The Start of Assistant A's Conversation with User|>

\#\#\# User:

\{question\_turn\_1\}
\par\medskip
\#\#\# Assistant A:

\{answer\_turn\_1\}\\
...
\par\medskip
\#\#\# User:

\{question\_turn\_last\}
\par\medskip
\#\#\# Assistant A:

\{answer\_turn\_last\}
\par\medskip
\#\#\# Document:

\{passages\}
\par\medskip
<|The End of Assistant A's Conversation with User|>
}}
\caption{\small Prompt used for the \mtb\ judge}
\label{fig:mtbench-prompt}
\end{figure}

\subsection{IDK correction to the metrics}
\label{sec:idk-judge}

To determine whether a model response corresponds to an IDK (I-Don't-know) answer, we use an LLM judge that we developed for this purpose. The judge sends the prompt of Figure~\ref{fig:idk-prompt} to GPT-4o-mini (2024-07-18). To measure the judge's performance, we evaluated its predictions on the human evaluation results across model responses and the reference answer, where the judge achieved over 97\% accuracy. We assess the impact of the IDK judge on the unanswerable questions using our reference answers and responses from frontier models (GPT-4o, Llama 3.1 405B Instruct) in Table~\ref{tab:idk_judge_quality}. These observations show that the IDK judge is needed to correct the scores. 

\begin{figure}[ht]
\noindent\fbox{%
    \parbox{\columnwidth-0.25cm}{%
\scriptsize
Read an inquiry and a response to it.  Decide if the response contains, in part or in whole, an indication that there is not sufficient information to fully answer the inquiry.  Use only one of the following labels: yes, no, partial.  Answer "yes" if the response indicates that the inquiry cannot be answered, for example, due to lack of information (in a reference document or text).  Answer "partial" if the response indicates insufficient information (in a reference document or text) to answer part of the inquiry, but still attempts to answer other part of the inquiry that can be answered.  Answer "no" if the response does not indicate a lack of information preventing answering part of the inquiry, even if some reasoning steps or general knowledge are needed to address the inquiry.
\par\medskip
[Inquiry]

How can I learn SQL without taking a course?
\par\medskip
[Response]

I’m unable to provide an answer regarding how you can learn SQL without taking a course. However, I can assist with the No SQL databases topic.
\par\medskip
[Output]

yes
\par\medskip
\emph{[Additional ICL examples]}
\par\medskip
[Inquiry]

\{inquiry\}
\par\medskip
[Response]

\{response\}
\par\medskip
[Output]

}}
\caption{\small IDK judge prompt}
\label{fig:idk-prompt}
\end{figure}

\begin{table}[!htbp]
    \centering\small
   \setlength{\tabcolsep}{2.5pt}
\begin{tabular}{l|rr|rr|rr}
\toprule
Model	&	\ragasF	&	\underline{\ragasF}	&	\radbench	&	\underline{\radbench}	&	\agg	&	\underline{\agg}	\\
\midrule
Reference	&	0.18	&	0.87	&	0.89	&	0.87	&	0.75	&	0.87	\\
GPT-4o	&	0.19	&	0.20	&	0.66	&	0.20	&	0.26	&	0.20	\\
Llama 3.1	&	0.21	&	0.20	&	0.69	&	0.20	&	0.27	&	0.20	\\
\bottomrule
\end{tabular}
    \caption{\small Generation results in the Reference ($\bullet$) retrieval setting before and \underline{after conditioning} with IDK judge on 55 unanswerable questions}
    \label{tab:idk_judge_quality}
\end{table}

\section{Synthetic conversation details}

We next provide additional details on the \benchmarksynth\ benchmark of synthetic multi-turn conversations. 

\subsection{Synthetic conversation generation}
\label{sec:synthetic-implementation-details}

To create \benchmarksynth, we extended the recently proposed framework of \cite{leegunasekara2024} to automatically generate user questions corresponding to the question types used in \benchmark. To classify question types within the synthetic conversations, we employed a
question type classifier trained on metadata derived
from the human-generated conversations in \benchmark. Finally, we used Mixtral 8x22B Instruct v0.1 as the LLM for the conversation generation and restricted the conversations to a maximum of $8$ turns, as further increasing the number of turns tends to lead to repetitive user questions and a higher likelihood of hallucinated agent responses.

\begin{table*}[t]
\begin{subtable}[t]{\columnwidth}
\centering
\small
\begin{tabular}{lcc}
\toprule
& {\bf \benchmark} & {\bf \benchmarksynth} \\
\midrule
Command-R+ (104B) & 0.76 & 0.83 \\
GPT-4o-mini & 0.71 & 0.81 \\
Llama 3.1 405B Instruct & 0.75 & 0.85 \\
Mixtral 8x22B Instruct & 0.61 & 0.79 \\
Qwen 2.5 (72B) & 0.72 & 0.82 \\
\bottomrule
\end{tabular}
\caption{\small Based on \ragasF}
\label{tbl:ragasf_results_human_vs_synthetic}
\end{subtable}
\hfill
\begin{subtable}[t]{\columnwidth}
\centering
\small
\begin{tabular}{lcc}
\toprule
& {\bf \benchmark} & {\bf \benchmarksynth} \\
\midrule
Command-R+ (104B) & 0.33 & 0.38 \\
GPT-4o-mini & 0.27 & 0.34 \\
Llama 3.1 405B Instruct & 0.33 & 0.4 \\
Mixtral 8x22B Instruct & 0.29 & 0.41 \\
Qwen 2.5 (72B) & 0.3 & 0.38 \\
\bottomrule
\end{tabular}
\caption{\small Based on BERT-K-Prec}
\label{tbl:berkprec_results_human_vs_synthetic}
\end{subtable}
\caption{\small Comparing faithfulness of models on the human-generated \benchmark\ and synthetic \benchmarksynth\ using two metrics: \ragasF\ and BERT-K-Prec}
\end{table*}

\begin{table*}[!thb]
\small
\centering
\begin{tabular}{c||l|l|l|l|l||l|l|l|l}
\toprule
&	& \multicolumn{4}{c||}{\bf Recall}			&	\multicolumn{4}{c}{\bf nDCG} \\
	&	&	@1	&	@3	&	@5	&	@10	&	@1	&	@3	&	@5	&	@10	\\
 \midrule
\multirow{ 2}{*}{\clapnq\ (208)}	&	Last Turn	&	0.20	&	0.43	&	0.53	&	0.65	&	0.48	&	0.46	&	0.49	&	0.54	\\
	&	Query Rewrite	&	0.22	&	0.45	&	0.56	&	0.70	&	0.54	&	0.50	&	0.53	&	0.59	\\
 \midrule
\multirow{ 2}{*}{\fiqa\ (180)}	&	Last Turn	&	0.15	&	0.34	&	0.44	&	0.55	&	0.38	&	0.36	&	0.40	&	0.45	\\
	&	Query Rewrite	&	0.18	&	0.39	&	0.50	&	0.63	&	0.43	&	0.41	&	0.46	&	0.52	\\
 \midrule
\multirow{ 2}{*}{\govt\ (203)}	&	Last Turn	&	0.18	&	0.42	&	0.50	&	0.58	&	0.42	&	0.44	&	0.46	&	0.49	\\
	&	Query Rewrite	&	0.21	&	0.47	&	0.56	&	0.67	&	0.47	&	0.48	&	0.51	&	0.56	\\
 \midrule
\multirow{ 2}{*}{\cloud\ (189)}	&	Last Turn	&	0.19	&	0.38	&	0.48	&	0.56	&	0.40	&	0.39	&	0.42	&	0.47	\\
	&	Query Rewrite	&	0.20	&	0.40	&	0.47	&	0.57	&	0.42	&	0.41	&	0.43	&	0.48	\\
 \midrule
\multirow{ 2}{*}{All (780)}	&	Last Turn	&	0.18	&	0.39	&	0.49	&	0.58	&	0.42	&	0.41	&	0.45	&	0.49	\\
	&	Query Rewrite	&	0.20	&	0.43	&	0.52	&	0.64	&	0.46	&	0.45	&	0.48	&	0.54	\\
\bottomrule
\end{tabular}
\caption{\small Elser Retrieval Performance of models on our benchmark using Recall and nDCG metrics per domain}
\label{tab:retrieval-results-domain}
\end{table*}

\newpage

\subsection{Faithfulness of models on synthetic data vs human-generated data}
\label{sec:faithfulness-synthetic}

We next provide detailed evaluation results on the comparison of faithfulness of models on \benchmark\ and \benchmarksynth, discussed in Section \ref{sec:synthetic}. Tables \ref{tbl:ragasf_results_human_vs_synthetic} and \ref{tbl:berkprec_results_human_vs_synthetic} show faithfulness on the two benchmarks based on the \ragasF\ and BERT-K-Prec metrics, respectively. Using either metric, we see that models across the board receive a higher faithfulness score on the synthetic than on the human-generated data. 

\section{Detailed retrieval results}
\label{sec:retrieval-tables}

Expanding on Table~\ref{tab:retrieval-results}, which shows average retrieval results, in Table~\ref{tab:retrieval-results-domain} we provide detailed retrieval results per domain for Elser, the best performing retriever.

\section{Detailed generation results}
\label{sec:generation-tables}

\begin{table*}[!htbp]
\begin{subtable}[t]{\textwidth}
\centering
\small
\setlength{\tabcolsep}{2.5pt}
\begin{tabular}{l|rrr|rrr|rrr|r}
\toprule
& \multicolumn{3}{c}{Overall} & \multicolumn{3}{c}{Answerable} & \multicolumn{3}{c}{Partial} & Unans. \\
& \ragasF & \radbench & \agg & \ragasF & \radbench & \agg & \ragasF & \radbench & \agg & 
\\
\midrule
Reference & 0.87 & 0.95 & 0.88 & 0.88 & 0.96 & 0.88 & 0.71 & 0.88 & 0.83 & 0.87 
\\
\midrule
Command-R+ (104B) & \textbf{0.76} & 0.69 & 0.44 & \textbf{0.82} & 0.74 & \underline{0.47} & \textbf{0.59} & 0.63 & 0.36 & 0.13 \\
GPT-4o & \underline{0.75} & \textbf{0.76} & \underline{0.45} & \textbf{0.82} & \textbf{0.81} & \textbf{0.48} & 0.53 & \underline{0.71} & 0.35 & 0.20 \\
GPT-4o-mini & 0.71 & \underline{0.75} & 0.43 & 0.77 & 0.79 & 0.44 & 0.39 & 0.62 & 0.30 & \underline{0.34} \\
Llama 3.1 405B Instruct & \underline{0.75} & 0.74 & \textbf{0.47} & \underline{0.81} & 0.79 & 0.50 & \underline{0.58} & 0.66 & \underline{0.37} & 0.20 \\
Llama 3.1 70B Instruct & 0.69 & 0.66 & 0.44 & 0.74 & 0.69 & 0.45 & 0.42 & 0.47 & 0.27 & \textbf{0.44} \\
Llama 3.1 8B Instruct & 0.55 & 0.59 & 0.36 & 0.59 & 0.62 & 0.38 & 0.34 & 0.47 & 0.24 & 0.33 \\
Mixtral 8x22B Instruct & 0.61 & 0.69 & 0.41 & 0.68 & 0.75 & 0.45 & 0.41 & 0.68 & 0.33 & 0.00 \\
Qwen 2.5 (72B) & 0.72 & 0.74 & 0.44 & 0.79 & \underline{0.80} & \underline{0.47} & 0.53 & \textbf{0.72} & \textbf{0.38} & 0.07 \\
Qwen 2.5 (7B) & 0.68 & 0.72 & 0.43 & 0.74 & 0.77 & 0.46 & 0.44 & 0.67 & 0.36 & 0.11 \\
\bottomrule
\end{tabular}
\caption{\small By question answerablity}
\label{tbl:judge_results_gold_by_answerability}
\end{subtable}
\par\bigskip
\begin{subtable}[t]{\textwidth}
\centering
\small
\begin{tabular}{l|rr|rr|rr}
\toprule
& \multicolumn{2}{c}{\ragasF} & \multicolumn{2}{c}{\radbench} & \multicolumn{2}{c}{\agg} \\
& TURN 1 & > TURN 1 & TURN 1 & > TURN 1 & TURN 1 & > TURN 1 \\
\midrule
Reference & 0.89 & 0.86 & 0.97 & 0.95 & 0.89 & 0.87 \\
\midrule
Command-R+ (104B) & 0.83 & \textbf{0.75} & 0.70 & 0.69 & 0.46 & 0.43 \\
GPT-4o & \textbf{0.86} & \underline{0.74} & \underline{0.78} & \textbf{0.76} & \textbf{0.54} & \underline{0.44} \\
GPT-4o-mini & \underline{0.84} & 0.69 & \textbf{0.79} & \underline{0.74} & \underline{0.50} & 0.42 \\
Llama 3.1 405B Instruct & 0.81 & \underline{0.74} & 0.74 & \underline{0.74} &\underline{0.50} & \textbf{0.46} \\
Llama 3.1 70B Instruct & 0.80 & 0.68 & 0.69 & 0.66 & \underline{0.50} & 0.42 \\
Llama 3.1 8B Instruct & 0.66 & 0.53 & 0.56 & 0.59 & 0.41 & 0.36 \\
Mixtral 8x22B Instruct & 0.82 & 0.58 & 0.73 & 0.69 & 0.47 & 0.40 \\
Qwen 2.5 (72B) & \underline{0.84} & 0.70 & 0.77 & \underline{0.74} & 0.51 & 0.43 \\
Qwen 2.5 (7B) & 0.82 & 0.65 & 0.71 & 0.72 & 0.48 & 0.42 \\
\bottomrule
\end{tabular}
\caption{\small By first vs subsequent turns}
\label{tbl:judge_results_gold_by_turns}
\end{subtable}
\par\bigskip
\begin{subtable}[t]{\textwidth}
\centering
\small
\setlength{\tabcolsep}{2.3pt}
\begin{tabular}{l|cccc|cccc|cccc}
\toprule
& \multicolumn{4}{c}{\ragasF} & \multicolumn{4}{c}{\radbench} & \multicolumn{4}{c}{\agg} \\
& \clapnq\ & \fiqa & \govt & \cloud & \clapnq\ & \fiqa & \govt & \cloud & \clapnq\ & \fiqa & \govt & \cloud \\
\midrule
Reference & 0.86 & 0.89 & 0.85 & 0.87 & 0.96 & 0.95 & 0.95 & 0.95 & 0.88 & 0.88 & 0.88 & 0.87 \\
\midrule
Command-R+ & \textbf{0.78} & \textbf{0.76} & \underline{0.76} & 0.73 & 0.72 & 0.70 & 0.68 & 0.66 & 0.47 & \underline{0.39} & 0.43 & 0.45 \\
GPT-4o & \underline{0.73} & \underline{0.74} & \textbf{0.77} & \textbf{0.78} & \textbf{0.77} & \textbf{0.74} & \textbf{0.78} & \textbf{0.74} & \textbf{0.48} & 0.38 & \underline{0.47} & 0.45 \\
GPT-4o-mini & 0.69 & 0.71 & 0.71 & 0.75 & \underline{0.76} & \textbf{0.74} & 0.75 & \underline{0.73} & 0.45 & 0.37 & 0.44 & 0.44 \\
Llama 3.1 405B Inst. & 0.72 & \textbf{0.76} & \textbf{0.77} & \underline{0.76} & 0.75 & \underline{0.73} & 0.75 & 0.72 & \textbf{0.48} & \textbf{0.41} & \textbf{0.49} & \textbf{0.49} \\
Llama 3.1 70B Inst. & 0.66 & 0.67 & 0.73 & 0.71 & 0.66 & 0.62 & 0.69 & 0.65 & 0.44 & 0.37 & 0.45 & \underline{0.47} \\
Llama 3.1 8B Inst. & 0.52 & 0.56 & 0.56 & 0.56 & 0.62 & 0.55 & 0.60 & 0.57 & 0.37 & 0.32 & 0.38 & 0.38 \\
Mixtral 8x22B Inst. & 0.60 & 0.60 & 0.61 & 0.64 & 0.72 & 0.68 & 0.70 & 0.68 & 0.45 & 0.34 & 0.42 & 0.44 \\
Qwen 2.5 (72B) & 0.68 & 0.70 & 0.74 & \underline{0.76} & 0.75 & \underline{0.73} & \underline{0.76} & \underline{0.73} & \underline{0.46} & 0.37 & 0.46 & 0.46 \\
Qwen 2.5 (7B) & 0.70 & 0.63 & 0.68 & 0.69 & 0.72 & 0.69 & 0.74 & 0.71 & \textbf{0.48} & 0.35 & 0.44 & 0.44 \\
\bottomrule
\end{tabular}
\caption{\small By domain}
\label{tbl:judge_results_gold_by_Domains}
\end{subtable}
\caption{\small Detailed generation results in the Reference ($\bullet$) retrieval setting using three metrics, \ragasF, \radbench, and \agg, on three different dimensions: (a) answerability, (b) turns, and (c) domains}
\end{table*}

Expanding on the generation experiment results of Section \ref{sec:generation} we provide two additional sets of results:\\

\subsection{Generation results for all three metrics}
\label{sec:detailed-gen-results-all-metrics}

Expanding on Figure \ref{fig:generation-breakdown}, which shows generation results in the Reference Generation ($\bullet$) setting using a single metric, \agg, we now show generation results for the same setting using all three metrics: \agg, \radbench, and \ragasF. Tables \ref{tbl:judge_results_gold_by_answerability}, \ref{tbl:judge_results_gold_by_turns}, and \ref{tbl:judge_results_gold_by_Domains} present the results broken down by answerability, turns, and domain, respectively.\\

\subsection{Generation results by question type and multi-turn type}
\label{sec:detailed-gen-results-by-qt-and-mt}

\begin{figure*}[t]
\begin{subfigure}[t]{\textwidth}
    \centering
    \includegraphics[height=5cm, width=\textwidth]{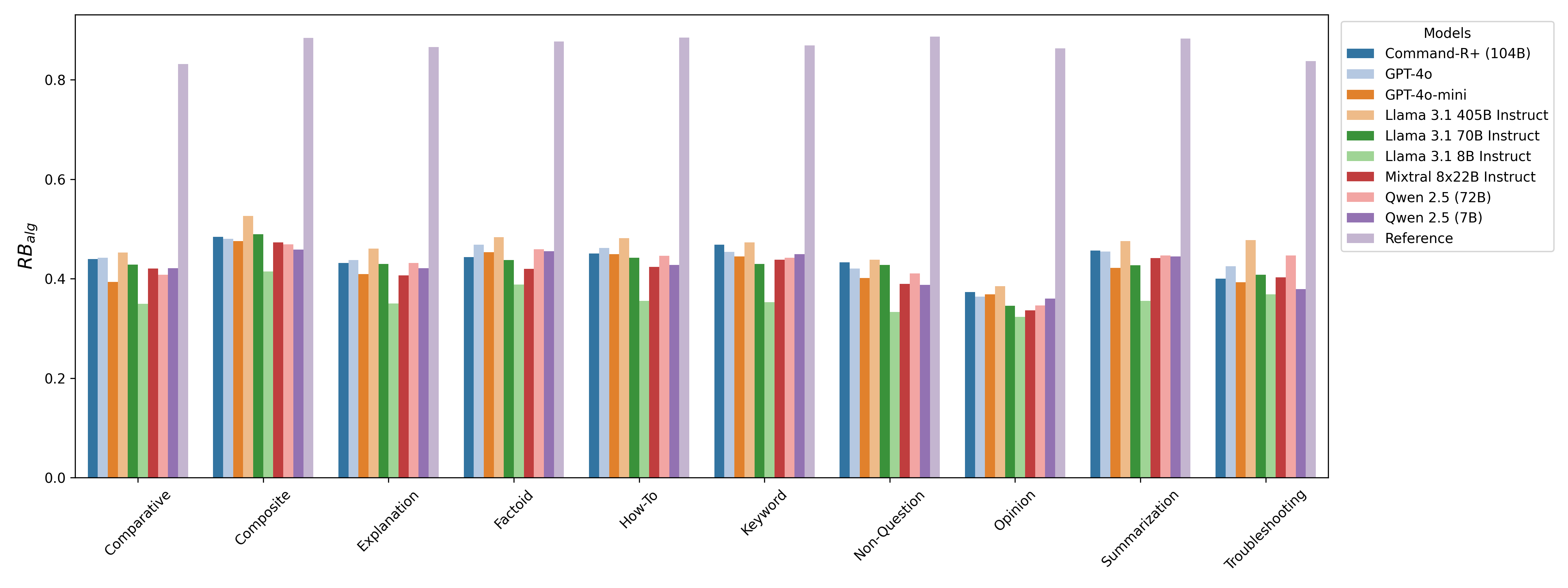}
    \vspace{-8mm}
    \caption{\small Based on \agg}
    \label{fig:model_performance_question_type_agg}
\end{subfigure}
\begin{subfigure}[t]{\textwidth}
    \centering
    \includegraphics[height=5cm, width=\textwidth]{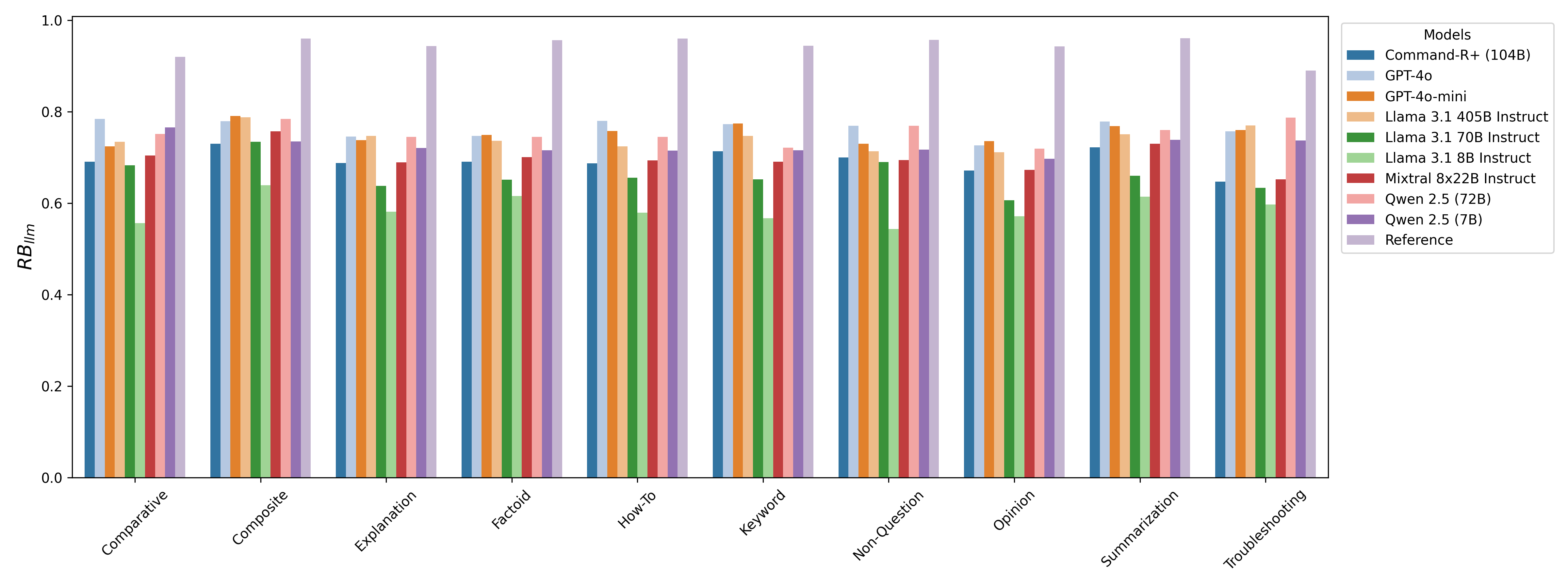}
    \vspace{-8mm}
    \caption{\small Based on \radbench}
    \label{fig:model_performance_question_type_radbench}
\end{subfigure}
\begin{subfigure}[t]{\textwidth}
    \centering
    \includegraphics[height=5cm, width=\textwidth]{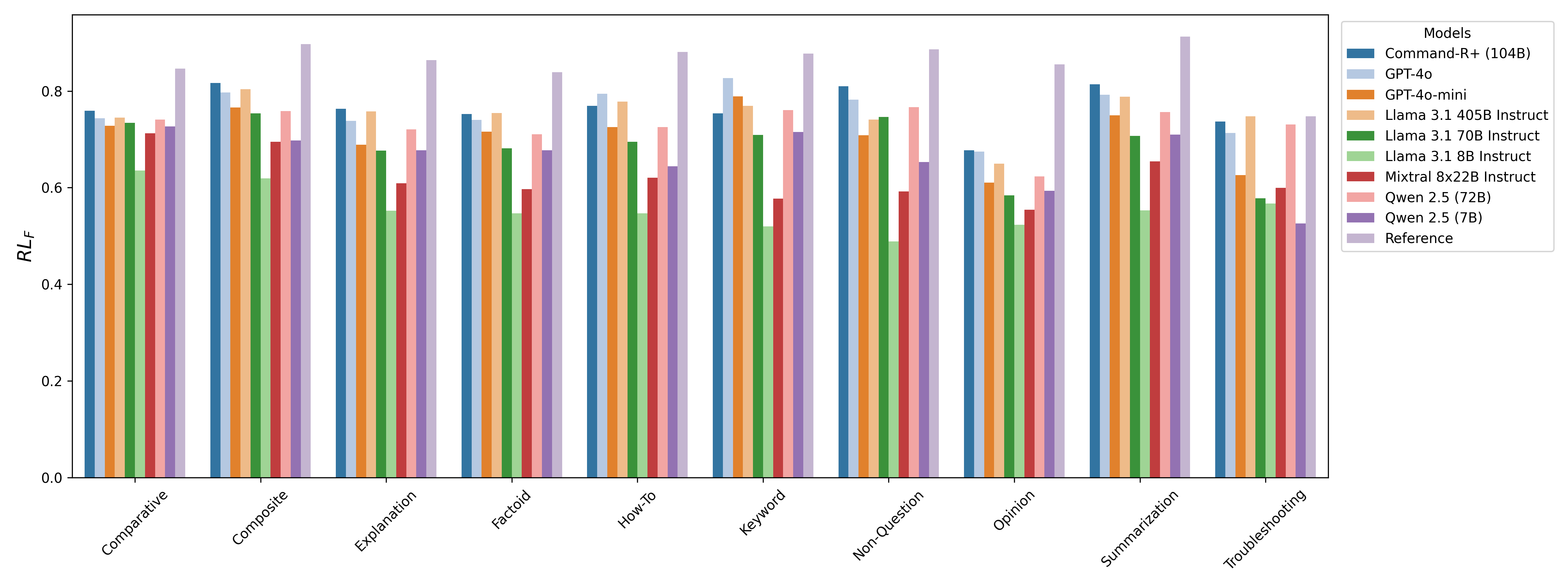}
    \vspace{-8mm}
    \caption{\small Based on \ragasF}
    \label{fig:model_performance_question_type_faithfulness}
\end{subfigure}
\vspace{-2mm}
\caption{\small Generation results in the Reference  ($\bullet$) retrieval setting by question type based on three metrics: \agg, \radbench, and \ragasF}
\label{fig:model_performance_qt_all}
\end{figure*}

\begin{figure*}[h]
    \centering
    \vspace{-5mm}
     \begin{subfigure}{0.27\textwidth}
        \centering
        \includegraphics[width=\textwidth]{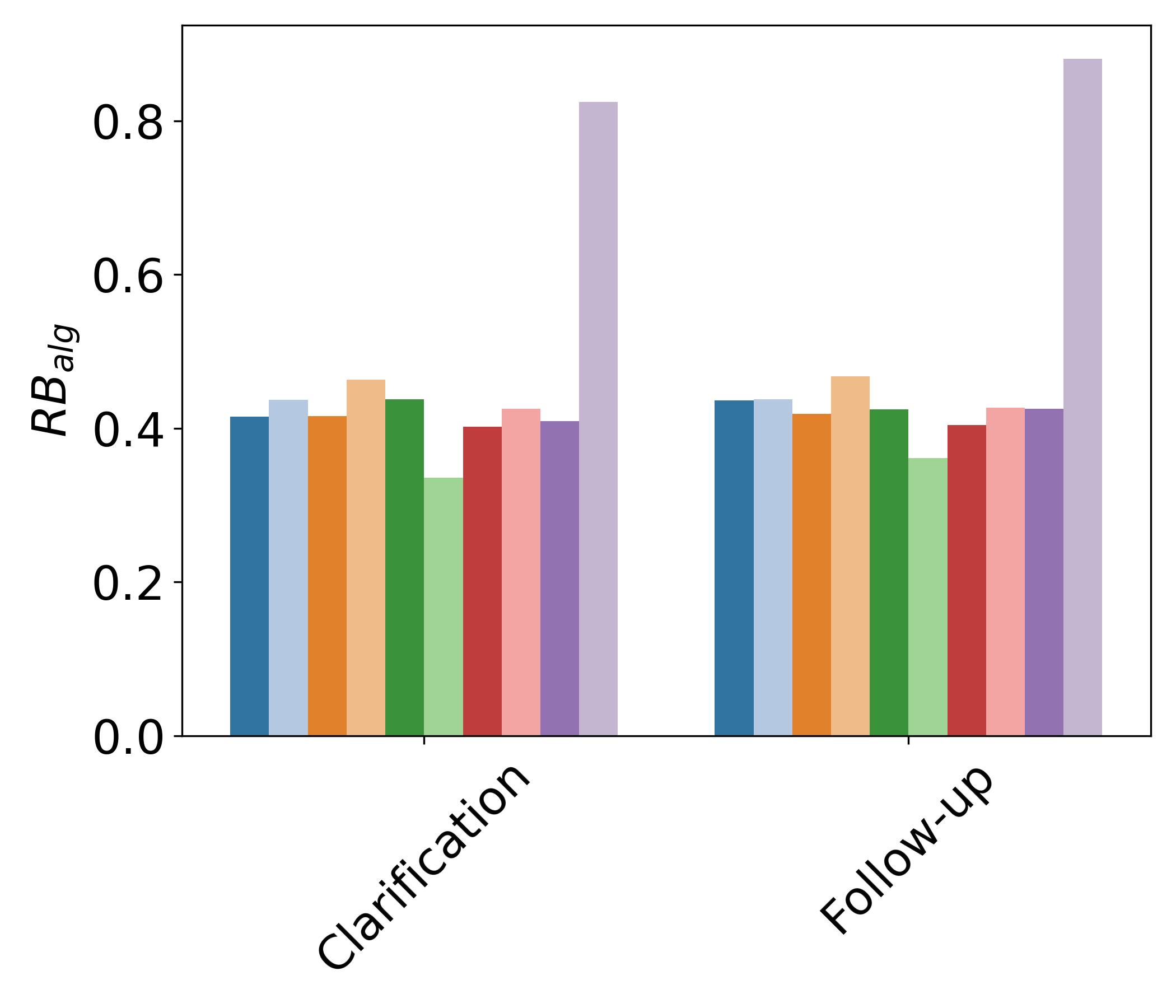}
        \vspace{-4mm}
        \caption{\small Based on \agg}
        \label{fig:model_performance_question_type_agg_mt}
    \end{subfigure}
    \hfill
     \begin{subfigure}{0.27\textwidth}
        \centering
    \includegraphics[width=\textwidth]{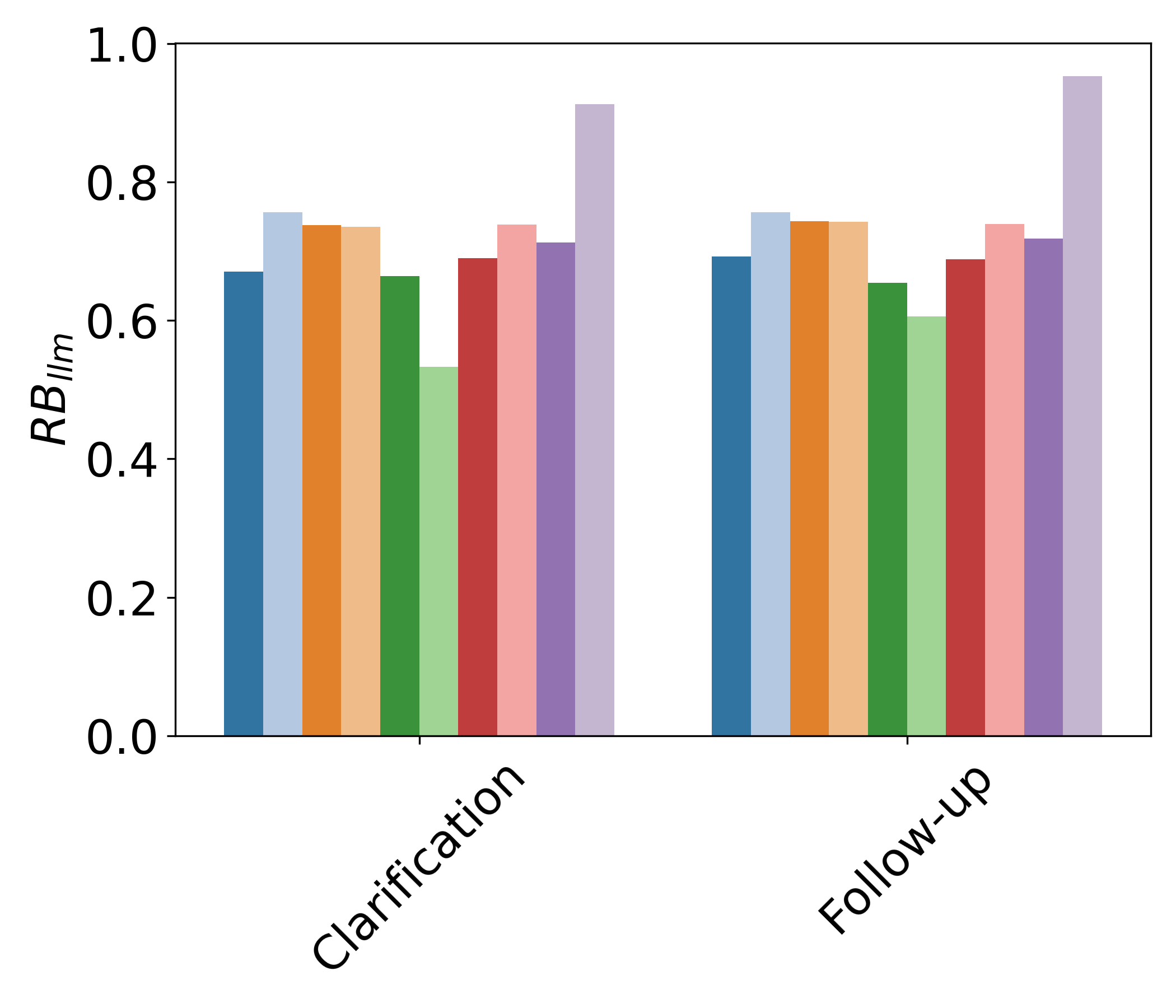}
        \vspace{-4mm}
        \caption{\small Based on \radbench}
        \label{fig:model_performance_question_type_radbench_mt}
    \end{subfigure}
    \hfill
     \begin{subfigure}{0.27\textwidth}
        \centering
        \includegraphics[width=\textwidth]{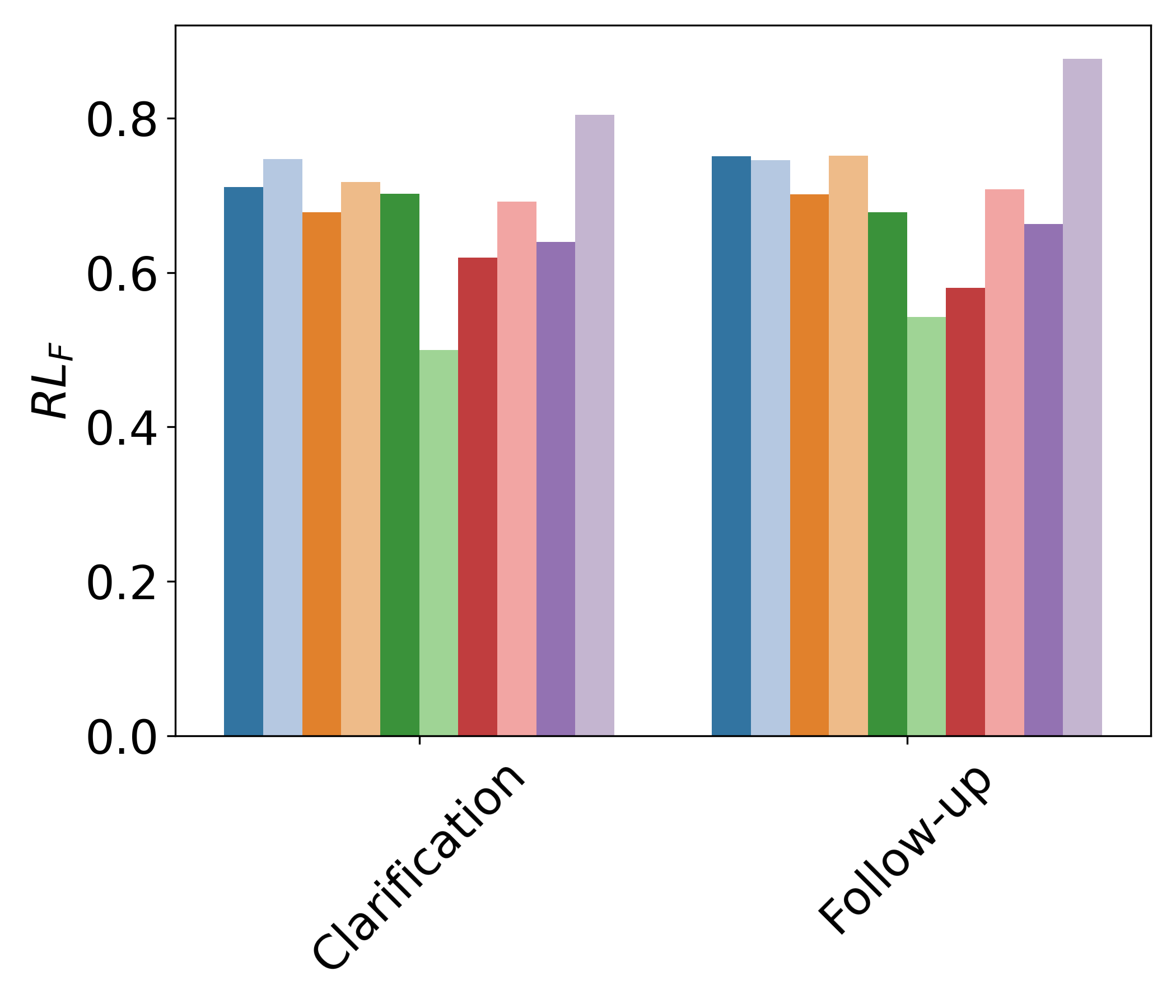}
        \vspace{-4mm}
        \caption{\small Based on \ragasF}
        \label{fig:model_performance_question_type_faithfulness_mt}
    \end{subfigure}
    \hfill
     \begin{subfigure}{0.17\textwidth}
        \centering
        \includegraphics[width=\textwidth]{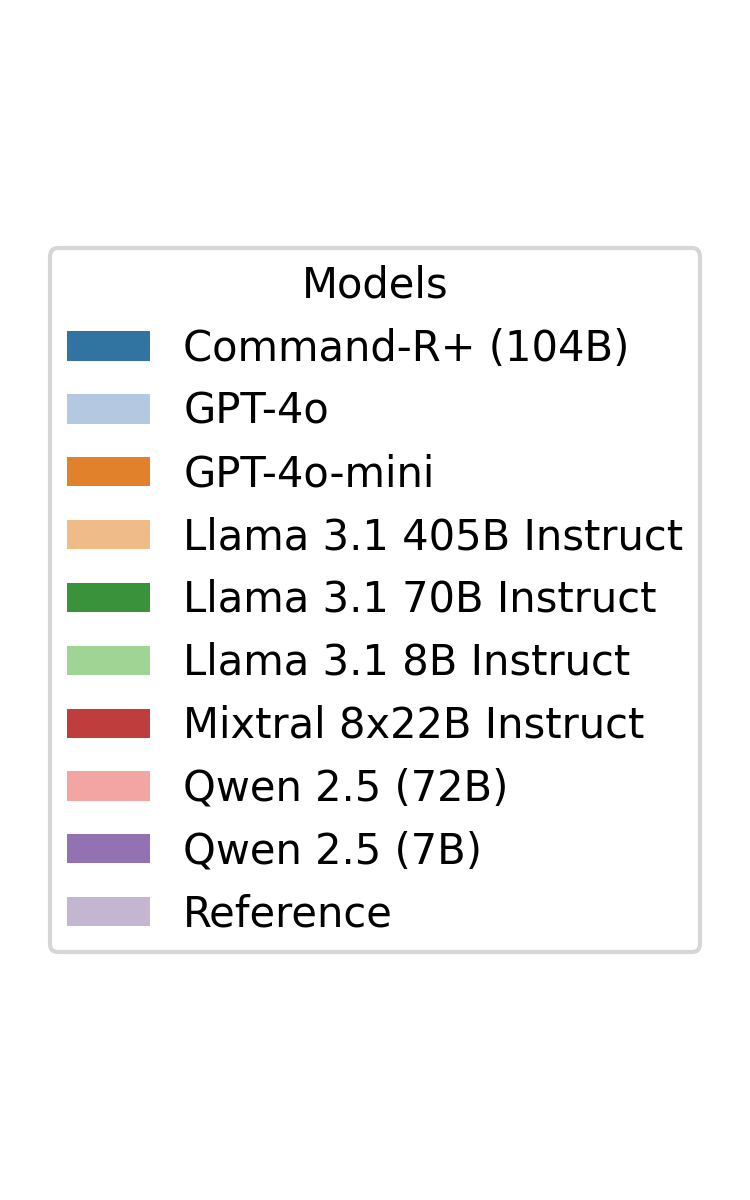}
        \label{fig:model_label}
    \end{subfigure}    
    \vspace{-4mm}
    \caption{\small Generation results in the Reference ($\bullet$) retrieval setting by multi-turn type based on three metrics: \agg, \radbench, and \ragasF}
    \label{fig:model_performance_mt_all}
\end{figure*}

Using the same three metrics, we also include generation results broken down by question and multi-turn type, shown in Figures \ref{fig:model_performance_qt_all} and \ref{fig:model_performance_mt_all}, respectively.

\end{document}